\documentclass[journal,compsoc, 10pts]{IEEEtran}

\ifCLASSINFOpdf%
\else
\fi

\usepackage{graphicx} 
\usepackage{epsfig} 
\usepackage{mathptmx} 
\usepackage{amsmath, bm} 
\usepackage{amssymb}  
\usepackage{wasysym}
\usepackage{float}
\usepackage{mathrsfs}
\usepackage{subfig,caption}
\usepackage{siunitx}
\usepackage{xspace}
\usepackage{xcolor, soul}
\usepackage{multicol}
\usepackage{multirow}
\usepackage{colortbl} 
\usepackage{booktabs}
\usepackage{rotating}
\usepackage{pdflscape}
\usepackage{siunitx}
\usepackage{tikz}
\usepackage{verbatim}
\usepackage{algorithm2e}
\usepackage{todonotes}
\usepackage[normalem]{ulem}
\usepackage{hyperref}
\usepackage{array}
\usepackage{mathrsfs}
\usepackage{tabularx}
\usepackage{soul}
\usepackage{ulem}
\usepackage{mathrsfs}
\usepackage{todonotes}
\usepackage{lscape}
\usepackage[export]{adjustbox}

\DeclareMathOperator{\Max}{Max}

\begin{document}
\newcommand\gridScene{G}

\newcommand\sceneXNum{n_{x}^{in}}
\newcommand\sceneYNum{n_{y}^{in}}
\newcommand\sceneZNum{n_{z}^{in}}

\newcommand\sceneXNumOut{n_{x}^{out}}
\newcommand\sceneYNumOut{n_{y}^{out}}
\newcommand\sceneZNumOut{n_{z}^{out}}

\newcommand\sceneXRes{s_{x}^{in}}
\newcommand\sceneYRes{s_{y}^{in}}
\newcommand\sceneZRes{s_{z}^{in}}

\newcommand\sceneXResOut{s_{x}^{out}}
\newcommand\sceneYResOut{s_{y}^{out}}
\newcommand\sceneZResOut{s_{z}^{out}}

\newcommand\sceneXMin{x_{min}}
\newcommand\sceneXMax{x_{max}}
\newcommand\sceneYMin{y_{min}}
\newcommand\sceneYMax{y_{max}}
\newcommand\sceneZMin{z_{min}}
\newcommand\sceneZMax{z_{max}}

\newcommand\factNum{n_{factors}}

\newcommand\factix{f_x^i}
\newcommand\factiy{f_y^i}
\newcommand\factiz{f_z^i}

\newcommand\voxelScene{v_{G}}
\newcommand\voxelObs{v_{H}}
\newcommand\voxelObsSet{V_{H}}

\newcommand\pointCloud{P}
\newcommand\ptsNum{N}
\newcommand\ptFeatNum{n_{feat}}

\newcommand\voxelNum{num_{voxels}}
\newcommand\voxelPts{num_{pts}}
\newcommand\ptFeats{num_{feats}^{pt}}

\newcommand\localVoxelPcFeats{num_{feats}^{voxel}}
\newcommand\localVoxelImgFeats{num_{feats}^{img}}

\newcommand\patchWidth{W_{patch}}
\newcommand\patchHeight{H_{patch}}

\newcommand\classNum{n_{classes}}
\newcommand\segThresh{thresh_{seg}}

\title{{R-AGNO-RPN}: A LIDAR-Camera Region Proposal Deep Network for
  Resolution-Agnostic Object Detection}
\author{Ruddy~Th{\'e}odose$^{1,2}$, Dieumet~Denis$^{1}$, Thierry
  Chateau$^{2}$, Vincent Fr{\'e}mont$^{3}$ and Paul
  Checchin$^{2}$
  \thanks{$^{1}$ Sherpa Engineering, R\&D Department, 333 Avenue
    Georges Clemenceau, 92000 Nanterre, France. {\tt\small
      [r.theodose, d.denis] @sherpa-eng.com}}%
  \thanks{$^{2}$ Universit\'e Clermont Auvergne, CNRS, SIGMA Clermont,
    Institut Pascal, F-63000 Clermont-Ferrand, France. {\tt\small
      firstName.lastName@uca.fr}} \thanks{$^{3}$ Centrale Nantes,
    LS2N, UMR CNRS 6004, Nantes, France. {\tt\small
      vincent.fremont@ls2n.fr}} }


\IEEEtitleabstractindextext{%
\begin{abstract}
  Current neural networks-based object detection approaches processing
  {LiDAR} point clouds are generally trained from one kind of {LiDAR}
  sensors. However, their performances decrease when they are tested
  with data coming from a different {LiDAR} sensor than the one used
  for training, i.e., with a different point cloud resolution. In this
  paper, {R-AGNO-RPN}, a region proposal network built on fusion of 3D
  point clouds and {RGB} images is proposed for 3D object detection
  regardless of point cloud resolution. As our approach is designed to
  be also applied on low point cloud resolutions, the proposed method
  focuses on object localization instead of estimating refined boxes
  on reduced data. The resilience to low-resolution point cloud is
  obtained through image features accurately mapped to Bird's Eye View
  and a specific data augmentation procedure that improves the
  contribution of the {RGB} images. To show the proposed network's
  ability to deal with different point clouds resolutions, experiments
  are conducted on both data coming from the {KITTI} 3D Object
  Detection and the {nuScenes} datasets. In addition, to assess its
  performances, our method is compared to {PointPillars}, a well-known
  3D detection network. Experimental results show that even on point
  cloud data reduced by $80\%$ of its original points, our method is still
  able to deliver relevant proposals localization.
\end{abstract}
%
}
\maketitle
\IEEEdisplaynontitleabstractindextext%
\IEEEpeerreviewmaketitle%
\section{Introduction}
Autonomous driving in urban scenes remains nowadays a tremendous
challenge. The autonomous vehicle must simultaneously manage multiple
aspects of perception such as scene analysis, traffic sign recognition
or moving object localization before deciding which action must be
engaged to preserve the safety of the other users and its own safety
too. 3D object detection for autonomous vehicles is an active research
topic since identifying the parts of the scene that can interfere with
the vehicle trajectory, is a crucial task within the navigation
pipeline. In this field, most advances rely essentially on {LiDAR}
sensors data.

Indeed, {LiDAR} sensors deliver 3D point cloud depicting the
surroundings of the sensor with high accuracy. However, the 3D point
clouds are not delivered in a unique and normalized data structure,
and the spatial positioning of points is irregular. Moreover, little
information on the surfaces' appearance is provided.  By contrast,
cameras deliver dense and structured data depending on which part of
the spectrum is captured (color information, infrared, etc.). These
features make them particularly adapted for classification or
identification tasks, even on small or noisy images. However, the loss
of the depth information due to the projection process and their
narrow field of view, makes difficult the full 3D reconstruction of
the scene geometry.  Over the past years, deep neural networks have
proven their efficiency in perception tasks by using
camera images. Most methods aim to extend these techniques to 3D point
clouds. However, due to the data type differences between images and
3D {LiDAR} information, the 3D point cloud data structure for neural
networks is still an open problem. Some methods for example transform
their sparse data into dense and structured grids through voxelization
that can then use operators initially designed for image
processing~\cite{yan2018second}. Other works exploit architectures
based on {PointNet}~\cite{qi2017pointnet, qi2017pointnet++} with the
aim to directly process the 3D point cloud by exploiting the geometric
relationships between points and their
neighborhood~\cite{Shi_2019_CVPR, Chen_2019_ICCV}.

In the field of Autonomous Vehicles, most of them are equipped with at
least these two types of sensors in order to benefit their respective
assets while counterbalancing the disadvantages of each one.  However,
few methods employ the data fusion of the two sensors due to their
different data structure format and the constraints that have to be
defined between the two sensors (required calibration, two pieces of
data to process, 3D and image space to manage, etc.). Currently, the
majority of 3D object detection methods are purely
{LiDAR}-based. However, some methods exploit the outputs of 2D
processing techniques to constraint the locations of the possible
locations. For example, the use of image semantic
segmentation~\cite{vora2019pointpainting} or 3D frustums generated by
2D image detection~\cite{qi2018frustum} are useful priors for the 3D
detection task.

While advances focused on the design of architectures that allow to
maximize accuracy, most of the existing models are trained
from a given set of {LiDAR} parameters and do not
generalize well if tested on another set. If solutions have been
developed for camera sensors (like domain
adaptation~\cite{wang2018deep}), to the best of our knowledge, almost
no work has analyzed the effects of 3D point clouds issued from
different {LiDAR} sensors on pre-trained systems. Therefore, we
believe that this topic is a major stake, as the acquisition and
annotation data process for 3D detection is expensive and
time-consuming. Moreover, the reuse of a pre-trained deep network that
is agnostic to the installed 3D {LiDAR}, might bring a potential boost
in the annotation process.

\begin{figure}[htb]
  \centering%
  \subfloat[Input 2D image with projected input point cloud.]{%
    \includegraphics[width=0.95\linewidth]{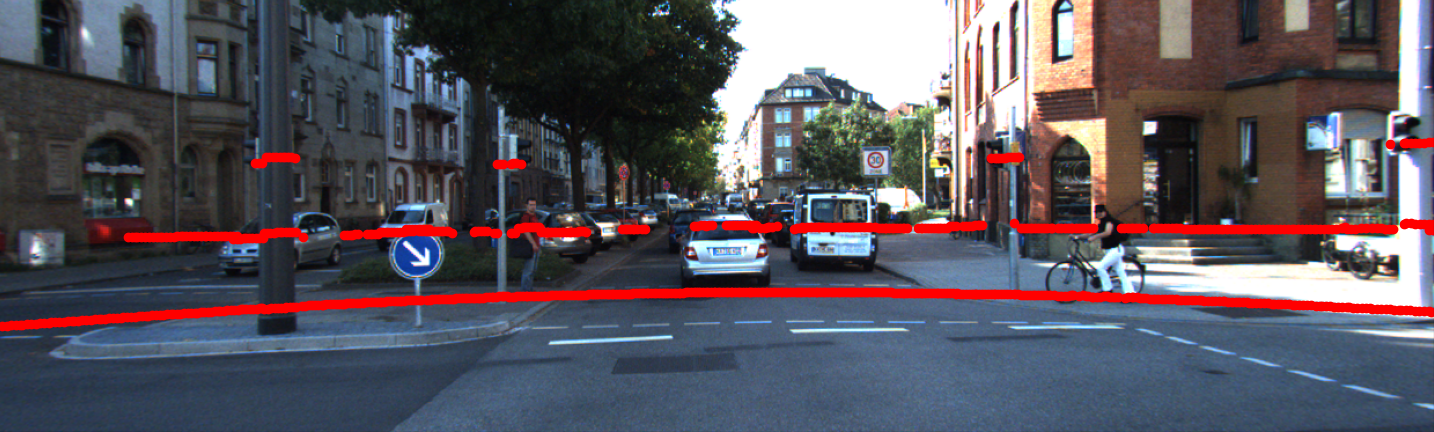}}
  \\
  \subfloat[Input 3D point cloud and predictions, the boxes are
  proposals returned by our network, the color of the boxes represents
  the classes and the score (red: \mbox{Car}, green:
  \mbox{Pedestrian}, cyan: \mbox{Cyclist})]{%
    \includegraphics[width=0.95\linewidth]{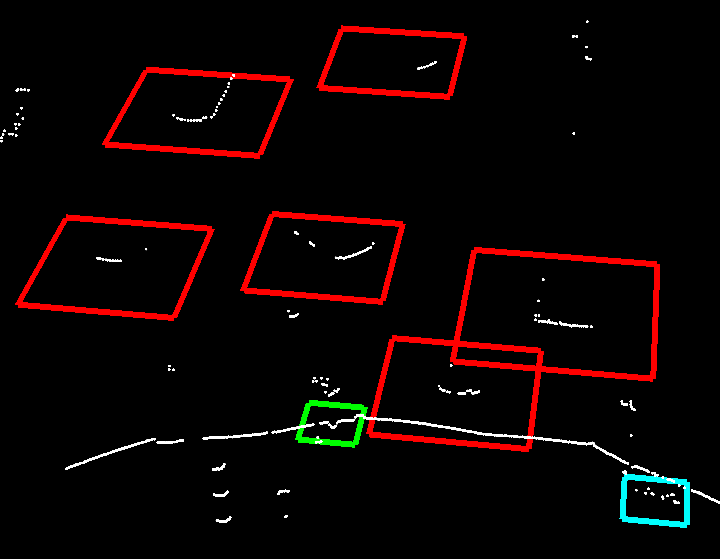}}
  
  \subfloat[Projections of proposals on camera image.]{%
    \includegraphics[width=0.95\linewidth]{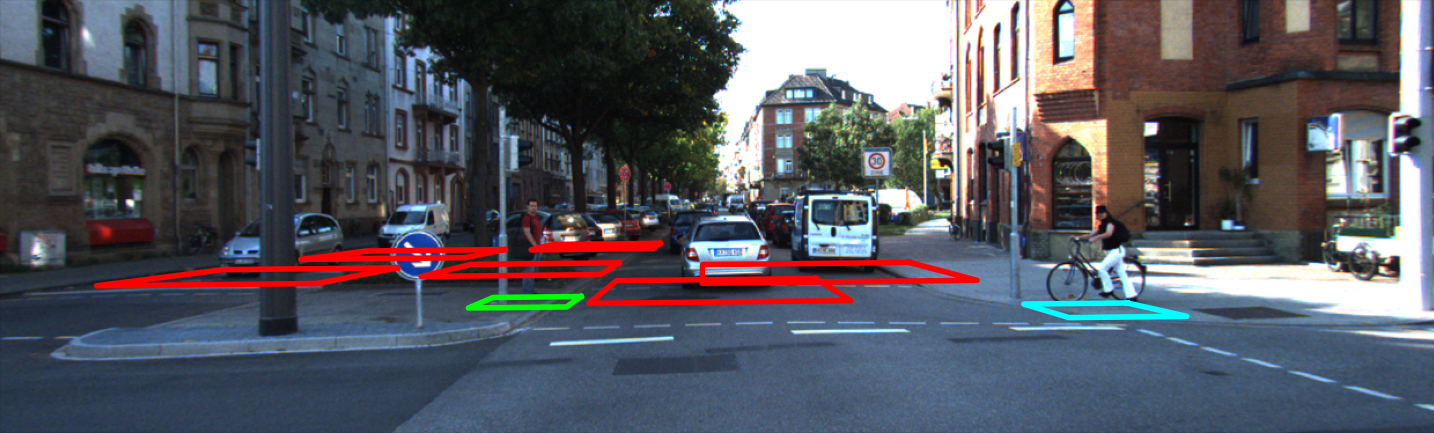}}
  \caption{Example of some results delivered by our network showing
    that correct proposals can be obtained even with low resolution
    point clouds.}%
  \label{fig:top_image} 
\end{figure}

In this paper, we propose a new Region Proposal Network ({RPN}) using
3D {LiDAR} data and camera images that is independent regarding the 3D
point cloud resolution as illustrated in
Figure~\ref{fig:top_image}. Our {RPN} first takes as input a color
image and returns a first feature map that is sampled on places
defined by the {LiDAR} 3D point clouds. These feature vectors can be
completed with local point clouds features re-projected into a Bird's
Eye View ({BEV}) map which is processed to return axis-aligned regions
of interest ({ROIs}) that encompass the potential obstacles. In fact,
it is difficult to estimate parameters such as object orientation if
the obstacle is hit by few laser
points. These proposals can then be employed for a subsequent system
whose role may be to track the potential obstacle in a sequence or
estimate finer parameters such as the orientation and the fitted
dimensions. This subsequent system will not be covered in this article
though.

As another contribution, we also propose a simple augmentation
procedure to make networks more robust to point cloud reduction,
resulting to a {RPN} that is able to adapt to different 3D point cloud
topologies.

To evaluate the performances of our system, the main tests are
performed on {KITTI} dataset~\cite{Geiger2013vision}. It is shown that
even if the number of 3D points seems to be low to realize some
correct predictions, our networks provides accurate regions of
interest as shown in Figure~\ref{fig:top_image}. Moreover, a more
difficult test is realized on the {nuScenes}
dataset~\cite{nuscenes2019} to illustrate how our network performs on
records with different data from different sensors ({LiDAR} and
camera). Related work on 3D object detection and cross {LiDAR} methods
are reviewed in Section~\ref{sec:related_work}. In
Section~\ref{sec:method}, we introduce the different elements that
compose our method from the input processing for the training to the
used architecture. Experiments and discussions are conducted in
Section~\ref{sec:experiments}.

\section{Related work}%
\label{sec:related_work}
In this section, some relevant networks aiming at 3D object detection
on point clouds and/or {RGB} images are firstly introduced. Secondly,
methods targeting on the management of point clouds at different
resolutions are then presented.

\subsection{3D Object Detection Methods}
3D object detection methods for autonomous driving can be divided into
three main groups depending on which types of data are used: methods
that infer 3D properties only from images, methods working exclusively
on point clouds and methods using both types of
data. Table~\ref{table:overview_detection_methods} summarizes these
approaches by recalling the raw input data used and the main paradigms
existing in the literature. Most of these approaches attempt to identify
targets categorized as \mbox{Car}, \mbox{Pedestrian} and
\mbox{Cyclist}. In fact, the {KITTI} dataset, whose benchmarks are
centered around those classes, is one of the oldest public dataset
allowing easier research on 3D object detection for autonomous
driving.  In the following paragraphs, those three groups of 3D
detection methods are analyzed.

\subsubsection{3D Object Detection using 2D Images}
Camera-based 3D object detection methods are generally less precise
than {LiDAR}-based ones due to the loss of depth information in the
image creation process. Hence, most of them rely on geometric
constraints priors. Here, the focus is given to monocular camera methods.

In~\cite{roddick2018orthographic}, a 3D grid is populated with
features from the input image by projecting voxels on the image,
knowing the camera parameters. To reduce the computational cost, the
resulting 3D grid is collapsed on the vertical axis.  The authors
of~\cite{shrivastava2020cubifae} train an encoder-decoder architecture
for depth estimation. The resulting intermediate features (after the
encoder, before the decoder) is then used for 3D object detection.
In~\cite{brazil2019m3d}, a two-branch architecture is defined, one
using regular convolutions to estimate global features, the other one,
called depth aware convolutions, splits the image into row blocks then
applies distinct kernels for each bin. The main assumption is that
each row block can be associated to a discretized depth due to
perspective effects (for example, lower rows often represent road and
close objects) and then a different operation for each depth bin is
applied.  In~\cite{Simonelli_2019_ICCV}, a 2D object detection method
is followed by a 3D object detection one. However, their loss function
splits the box parameters into groups to simplify the optimization
process.

\begin{table*}[!t]
  \caption{Summary of published networks aiming at 3D object detection
    from point clouds and/or {RGB} images. An overview of the state of
    experimental evaluation of methods useful for autonomous driving.}
  \scriptsize
  \begin{tabular}{p{1.5cm}p{1.7cm}p{2.5cm}p{6.6cm}p{1.8cm}p{1.6cm}}
    \toprule
    \textbf{Reference} & \textbf{Raw Input Data} & \textbf{Raw Data Processing} & \textbf{Overview} & \textbf{Object Types} & \textbf{Dataset{(s)} used} \\
    \midrule
    OFT~\cite{roddick2018orthographic} & {RGB} Image & - & Resnet Backbone, projection of 3D voxels on image feature maps, each 3D assigned with features inside the projection, then Voxelnet-like processing & \mbox{Car}s & {KITTI} \\
    M3D-RPN~\cite{brazil2019m3d} & {RGB} Image & - & DenseNet Backbone then 2 branches: (1) global feature extraction on whole image, (2) local ``Depth-Aware'' feature extractions.  Horizontally sliced image. Each slice is fed to its own convolution & \mbox{Car}, \mbox{Pedestrian}, \mbox{Cyclist} & {KITTI} \\
    MonoDIS~\cite{Simonelli_2019_ICCV} & {RGB} Image & - & ResNet Backbone, 2D and 3D heads, box parameters divided into groups for a better convergence & \mbox{Car}, \mbox{Pedestrian}, \mbox{Cyclist} & {KITTI} \\
    CubifAE-3D~\cite{shrivastava2020cubifae} & {RGB} Image & - & Multiple trainings: (1) depth estimation through encoder-decoder, (2) depth latent space for 3D detection & \mbox{Car}, \mbox{Pedestrian}, \mbox{Cyclist} and others & {KITTI}, {nuScenes}, {KITTI} Virtual 2 \\
    \midrule
    BirdNet~\cite{beltran2018birdnet} & Point Cloud & Discretization into {BEV} 3-channel image (height, density, reflectance) & Faster RCNN-like network & \mbox{Car}, \mbox{Pedestrian}, \mbox{Cyclist} & {KITTI} \\
    Complex-YOLO~\cite{simony2018complex} & Point Cloud & Discretization into {BEV} 3-channel image (height, density, reflectance) & YOLO-like network, complex angles for regression & \mbox{Car}, \mbox{Pedestrian}, \mbox{Cyclist} & {KITTI} \\
    VoxelNet~\cite{zhou2018voxelnet} & Point Cloud  & 3D voxelization & Learned voxel encoding for each voxel, 3D convolutions to flatten on the vertical axis then RPN & \mbox{Car}, \mbox{Pedestrian}, \mbox{Cyclist} & {KITTI} \\
    SECOND~\cite{yan2018second} & Point Cloud & 3D voxelization & Learned voxel Encoding, sparse 3D convolution to reduce the computational burden, RPN & \mbox{Car}, \mbox{Pedestrian}, \mbox{Cyclist} & {KITTI} ({nuScenes} thereafter) \\
    PointPillars~\cite{Lang_2019_CVPR} & Point Cloud & Column voxelization (no slices on vertical axis) & Learned voxel encoding, 2D convolutions, RPN & \mbox{Car}, \mbox{Pedestrian}, \mbox{Cyclist} & {KITTI}, {nuScenes} \\
    OHS~\cite{chen2019object} & Point Cloud & 3D voxelization, voxels represented by the mean content & Objects as sets of hotspots (non empty voxels belonging to an object), architecture inspired by SECOND or PointPillars according the dataset & \mbox{Car}, \mbox{Pedestrian}, \mbox{Cyclist} and others &  {KITTI}, {nuScenes} \\
    PointRCNN~\cite{Shi_2019_CVPR} & Point Cloud & - & Two-stage method: (1) 3D proposal generation through point cloud segmentation, each foreground point generating its own 3D proposal, (2) 3D box refinement & \mbox{Car}, \mbox{Pedestrian}, \mbox{Cyclist} & {KITTI} \\
    Fast Point R-CNN~\cite{Chen_2019_ICCV} & Point Cloud  & 3D voxelization & Two-stage method: (1) 3D region proposal through VoxelNet architecture, (2) pooling on VoxelNet's feature maps, fusion with the corresponding section of point cloud, refinement network on pooled and augmented point cloud & \mbox{Car}, \mbox{Pedestrian}, \mbox{Cyclist} & {KITTI} \\
    CenterPoint~\cite{yin2020center} & Point Cloud & 3D or column voxelization & Objects represented by their center on the classification map, architecture inspired by VoxelNet \& PointPillars & {nuScenes} classes & {nuScenes} \\
    PV-RCNN~\cite{shi2020pv} & Point Cloud & 3D Voxelization $+$ furthest point sampling to extract keypoints & Two-stage detection: (1) architecture inspired by SECOND for the region proposal, the 3D feature maps sampled at multi-scales using the computed keypoints, (2) RoI-Grid Pooling on the keypoints then Refinement Network & \mbox{Car}, \mbox{Cyclist} & {KITTI}\\
    Part-$A^{2}$ Net~\cite{shi2019part}  & Point Cloud & - & Two-stage
                                                             detection:
                                                             (1) point
                                                             cloud
                                                             semantic
                                                             segmentation,
                                                             part estimation for positive 3D points then 3D proposals, (2) point cloud pooling then refinement network to aggregate the estimated parts &  \mbox{Car}, \mbox{Pedestrian}, \mbox{Cyclist} & {KITTI} \\
    \midrule
    %
    %
    %
    MV3D~\cite{chen2017multi} & Point Cloud, {RGB} image & Discretization into {BEV} 3-channel image (height, density, reflectance) and Cylindrical projection & Two-stage detector: (1) {BEV} generates 3D proposals, (2) 3D proposals are projected on the {BEV}, the {RGB} image and on the cylindrical view, features from the 3 feature maps (each view) are pooled then merged to refine the boxes & Car & {KITTI} \\
    Frustum PointNet~\cite{qi2018frustum} & Point Cloud, {RGB} image & - & 2D detector on the image to create their 3D frustum, for each result, the 3D points inside the frustum used to estimate the corresponding 3D box with an architecture similar to PointNet++ & \mbox{Car}, \mbox{Pedestrian}, \mbox{Cyclist} & {KITTI} \\
    AVOD~\cite{ku2018joint} & Point Cloud, {RGB} & Discretization into {BEV} 3-channel image (height, density, reflectance) & Two-stage detector: (1) {BEV} and {RGB} images merged to generate 3D proposals (2) 3D proposals are projected on the {BEV} and the {RGB} image feature maps, the corresponding features are pooled then merged to refine the 3D boxes & \mbox{Car}, \mbox{Pedestrian}, \mbox{Cyclist} & {KITTI} \\
    PointFusion~\cite{xu2018pointfusion} & Point Cloud, {RGB} image & - & 2D detector to extract 2D crop and 3D frustum, 2D crop and corresponding point cloud processed by a ResNet and a PointNet, respectively. Features merged to estimate a refined 3D box & \mbox{Car}, \mbox{Pedestrian}, \mbox{Cyclist} & {KITTI}, SUN-RGBD \\
    ContFusion~\cite{liang2018deep} & Point Cloud, {RGB} image & - & 2D feature extraction, feature sampling through continuous convolution with the point cloud, reprojection of the sampled features on a {BEV} then 3D box prediction & Car & {KITTI}, TOR4D (private) \\
    IPOD~\cite{yang2018ipod} & Point Cloud, {RGB} image & 2D Semantic Segmentation & 3 parts: (1) using the 2D segmentation, all the 3D background points are discarded and proposals are generated from the remaining points, (2) a backbone network extracts local and global features from the whole point cloud,  (3) the features are extracted from the proposals and used in a refinement network & \mbox{Car}, \mbox{Pedestrian}, \mbox{Cyclist} & {KITTI} \\
    Frustum ConvNet~\cite{wang2019frustum} & Point Cloud, {RGB} image & - & 2D detector on the image to create their 3D frustum, for each result, the part of the point cloud inside the frustum is kept and voxelized along the frustum axis, convolutions are used along this axis to estimate the 3D box & \mbox{Car}, \mbox{Pedestrian}, \mbox{Cyclist} & {KITTI} \\
    PointPainting~\cite{vora2019pointpainting} & Point Cloud, {RGB} image & 2D Semantic Segmentation & Point features (3D coordinates, reflectance) concatenated with the score given by 2D segmentation. Experiments on multiple detectors (PointRCNN, PointPillars, etc.) & \mbox{Car}, \mbox{Pedestrian}, \mbox{Cyclist} & {KITTI}, {nuScenes} \\
    MVX-Net~\cite{sindagi2019mvx} & Point Cloud, {RGB} image & 3D Voxelization & VoxelNet used for the 3D detection and Backbone Faster RCNN for the 2D feature maps, 2 fusion methods tested: (1) 2D features sampled on point level and merged before voxel encoder, (2) 2D features sampled on voxel level and merged after voxel encoder & \mbox{Car}  & {KITTI} \\
    MMF~\cite{liang2019multi} & Point Cloud, {RGB} image & 3D Voxelization & Two-stage detector: (1) {RGB} image processed to generate intermediate feature maps and a depth map. These maps are merged with {LiDAR} features to generate 3D proposals. (2) {LiDAR} and {RGB} feature maps are pooled and merged to refine the estimated boxes & \mbox{Car} (main) & {KITTI}, TOR4D (private) \\
    \bottomrule
  \end{tabular}%
  \label{table:overview_detection_methods}
\end{table*}

\subsubsection{Point Cloud-based Methods}
These methods only use 3D point clouds provided by 3D {LiDAR}
sensors. This category can be divided into two subgroups: grid-based 
and point-based methods.

\textbf{Grid-based Methods}\\
The main achievements obtained in object detection occurred on
structured data, mainly {RGB} images. Hence, the main idea of
grid-based methods consists in turning the 3D point cloud into
structured data to allow the use of concepts and operators which were
successfully applied in 2D image processing such as convolutions. The
main drawback is that the precision depends on the discretization of
the grid. Some methods are based on existing 2D architectures to
extract detections from Bird Eye's View pseudo
images. {BirdNet}~\cite{beltran2018birdnet} and {Complex
  YOLO}~\cite{simony2018complex} are based on
{Faster-RCNN}~\cite{ren2017faster} and {YOLO}~\cite{redmon2016you,
  redmon2018yolov3}, respectively. Their
corresponding networks are applied on 3-channels
pseudo-image made by the concatenation of a height map, a reflectance
map and a density map.  {VoxelNet}~\cite{zhou2018voxelnet} was a major
milestone. In their paper, the authors describe an architecture
including encoding voxels features through PointNet-like methods. The
encoded grid is given to 3D dense convolutions and afterwards to 2D
convolutions in order to return the selected prior boxes and their
correction. However, the use of 3D dense convolutions on a large 3D
grid is slow and expensive. {SECOND}~\cite{yan2018second} rectifies
this by applying sparse convolutions~\cite{graham2017submanifold} on
the grid. {PointPillars}~\cite{Lang_2019_CVPR} extends the concept by
using directly columns in a 2.5D pseudo image instead of voxels in a
3D grid in order to speed up the
computation. {Voxel-FPN}~\cite{wang2019voxel}, inspired by the Feature
Pyramid Networks~\cite{lin2017feature}, uses voxel grids with
different resolutions to encode voxels at different
scales. Anchor-free methods, assimilating object detection to keypoint
detection~\cite{zhou2019objects, law2018cornernet} have also been
transposed to 3D object detection~\cite{chen2019object,
  yin2020center}. The main goal is not to change the input data
representation but the output format, as detections are no more
related to prior boxes.

\textbf{Point-based Methods}\\
Contrary to grid-based methods, point-based methods aim to deal at some
stage with raw point clouds, directly. This paradigm can be applied
for the whole method or jointly with grid representations. Moreover,
most of these methods are two-stage
detectors. {PointRCNN}~\cite{Shi_2019_CVPR} adopts a two-stage
approach: a first stage generates proposals from foreground point
segmentation, a second phase refines the proposals to estimate
targeted boxes. Fast {PointRCNN}~\cite{Chen_2019_ICCV} and
{PV-RCNN}~\cite{shi2020pv} employ a {VoxelNet}-like to generate
proposals that are processed by {PointNet}-like
networks. Part-$A^{2}$~\cite{shi2019part} directly estimates object
parts (top left, bottom right, etc.) from point cloud in the first
stage to improve the 3D box refinement stage.

\subsubsection{3D Object Detection using Multi-Sensor Fusion}
Contrary to the aforementioned methods that exploit only one sensor,
the ones presented in the following section focus on camera-{LiDAR}
data fusion to localize the surrounding objects. These
methods can be divided into two sub-groups. The first group makes use
of both sensor data within the same system. The second group rather
employs outputs from existing and pre-trained 2D methods as priors or
inputs for methods that focus on point clouds. Both groups are
introduced in the following paragraphs.

\textbf{Parallel Flows}\\
These methods generally use images as a second input to exploit
complementary nature of data.  {MV3D}~\cite{chen2017multi} and
{AVOD}~\cite{ku2018joint} belong to methods that simultaneously
process images and point clouds to extract intermediate
proposal. {MV3D}~\cite{chen2017multi} generates 3D proposals from a
{LiDAR} Bird Eye's View image and then, projects them onto the {LiDAR}
BEV, a {LiDAR} front view and an {RGB} image. Features inside the
projection of each modality are extracted through {ROI} pooling and
merged to estimate final boxes. {AVOD}~\cite{ku2018joint} extends the
idea by projecting all pre-built prior boxes on high-resolution
feature maps from {LiDAR} {BEV} and {RGB} images.
MVX-Net~\cite{sindagi2019mvx} add pooled features from {RGB} feature
maps into voxels that are used on a {VoxelNet}. The method described
in~\cite{liang2018deep} takes advantage of leveraging continuous
convolutions to merge feature maps from each sensor. The authors
of~\cite{liang2019multi} improve this architecture by adding related
tasks such as ground estimation and depth completion to boost the 3D
refinement sub-network.

\textbf{Sequential Streams}\\
2D processing systems have become more mature. Contrary to
above-mentioned, the methods introduced in this paragraph operate
sequentially and are 2D driven since they exploit results from {RGB}
processing as filters on point cloud. {Frustum
  PointNet}~\cite{qi2018frustum}, {Frustum
  Convnets}~\cite{wang2019frustum} and
{PointFusion}~\cite{xu2018pointfusion} exploit results from a 2D
detector to extract points that fall inside each detection and then
reduce the search space for each object detected by the image
detector. The frustums are then processed to precisely localize
objects. {IPOD}~\cite{yang2018ipod} removes the background 3D points
by exploiting a segmentation map from {RGB} camera
images. Each foreground point is used as a location for
prior boxes and then processed
afterwards. In~\cite{vora2019pointpainting}, the authors show the
performances improvement on several {LiDAR} detectors only by
integrating the semantic information from a segmentation map on the
input.

\subsection{Point Cloud-based Resolution-Agnostic Deep Learning}
To the best of our knowledge, even if domain adaptation has become an
active research field on images, few works on domain adaptation or
portability have been done on 3D point clouds, especially on outdoor
sensor data. As illustrated in~\cite{lambert2020performance}, each 3D
{LiDAR} has its own features about range, point distribution, data
coherency on disturbed conditions. The authors
of~\cite{piewak2019analyzing} perform a deep analysis of the
performance of their architecture for semantic segmentation of point
clouds from a 32-layer and a 128-layer {LiDAR}. 3D Domain Adaptation
is especially studied and analyzed on 3D CAD (Computer-Aided Design)
objects datasets such as {ModelNet}~\cite{wu20153d} and dense indoor
point clouds. Studies on open scenes are less usual and most of them
targets the semantic segmentation of point
clouds. In~\cite{achituve2020self}, a shared representation is learned
in a self-supervised fashion through the reconstruction of deformed 3D
point clouds. {PointDAN}~\cite{qin2019pointdan} designed an
architecture to learn cross domain local and global features to align
objects from two distinct distributions. Some methods such
as~\cite{wu2019squeezesegv2} employ Generative Adversarial Networks
(GAN) to close the gap between synthetic data and real-world
data. In~\cite{yi2020complete}, the authors propose a point cloud
completion method by assimilating the task as the reconstruction of
the underlying surface. A semantic segmentation network is then
trained from the reconstructed surface that serves as the
new canonical domain.

Our method does not directly aim at {LiDAR} domain adaptation.
Instead, the use of {RGB} data and our augmentation procedure enforce
some aspects of the network that makes it more resilient to point
cloud resolution variations.

\section{LiDAR-Camera Data Fusion Region Proposal Network Regardless
  of Point Cloud Resolution}%
\label{sec:method}
In this paper, we propose an end-to-end trainable method for a region
proposal network detection. The Region Proposal Network ({RPN}) takes
{RGB} images and point cloud as inputs. The objective
is to deliver a Bird's Eye View ({BEV}) Axis-Aligned Bounding Boxes
({AABB}) for the objects inside the search space that can be used in a
subsequent refinement network in order to accurately regress the
Oriented Bounding Boxes ({OBB}).  As raw images are also available,
and contrary to many two-stage methods, a semantic classification is
performed on proposals.  In this section, the chosen inputs
jointly with the architecture of the network and
the output generation procedure are described.

\subsection{{LiDAR} Data Preprocessing}
This section presents the set of operations applied on the 3D {LiDAR}
point cloud to produce a voxel-based representation. The voxelization
paradigm introduced in~\cite{Lang_2019_CVPR} is used. The
point cloud is turned into columns voxels with no vertical
discretization compared to a 3D voxel grid. This way, a {BEV} grid/map
whose each non-zero pixel is assigned to a non-empty voxel is
generated. The terms ``pixel'' and ``voxel'' are used interchangeably
when referred to the {BEV} image.

The $x$-axis is oriented forwards, the $y$-axis is oriented to the
left and the $z$-axis is oriented upwards.  A 3D point cloud is
defined as a set
$\mathbf{P} = {\{ {[x_{i}, y_{i}, z_{i}]}^{T} \in \mathbb{R}^{3} \}}_{i
  \in 1 \ldots \ptsNum}$ with $(x_{i}, y_{i}, z_{i})$ the location of
the $i$-th point in the 3D space.  The search space is restricted to
the interval $[\sceneXMin, \sceneXMax]$ on the $x$ axis and
$[\sceneYMin, \sceneYMax]$ on the $y$ axis. This point cloud is
discretized and grouped into column voxels of size
$(\sceneXRes, \sceneYRes)$. The resulting {BEV} grid contains
$\sceneXNum \times \sceneYNum$ voxels where
\begin{equation*}
  \sceneXNum = \left\lfloor \frac{\sceneXMax-\sceneXMin}{\sceneXRes}
  \right\rfloor, \
  \sceneYNum = \left\lfloor \frac{\sceneYMax-\sceneYMin}{\sceneYRes} \right\rfloor. 
\end{equation*}
The {LiDAR} points are grouped according to the voxel they belong to.
The number of points in a voxel is variable and depends on many
parameters such as the topology of the scene, the distance of the
voxel from the sensor, the used sensor, and the pose of the {LiDAR}
towards the scene. For each non-empty voxel, a Normal Distribution
Transform ({NDT})~\cite{biber2003normal} is operated on the points
inside the voxel.  The Normal Distribution Transform is a simple
method that turns a set of points into a multivariate Gaussian
function. For 3D points, a voxel is then represented by a feature
voxel which is at least a vector of size $9$ composed by:
\begin{itemize}
\item the mean $\bm{\mu} = {(\mu_{x}, \mu_{y}, \mu_{z})}^{T} \in \mathbb{R}^{3}$;
\item the symmetric covariance matrix $\bm{\Sigma} \in \mathbb{R}^{3 \times 3}$.
\end{itemize}
The six extreme coordinates of the 3D points inside the voxel are also
included in the feature voxel. Therefore, the 3D information
representing the voxel is encoded as a vector of $3+6+6=15$
dimensions.

\begin{figure}[!t] 
  \centering \subfloat[Outputs of voxelization, each large point is a
  main point.]{%
    \includegraphics[width=0.95\linewidth]{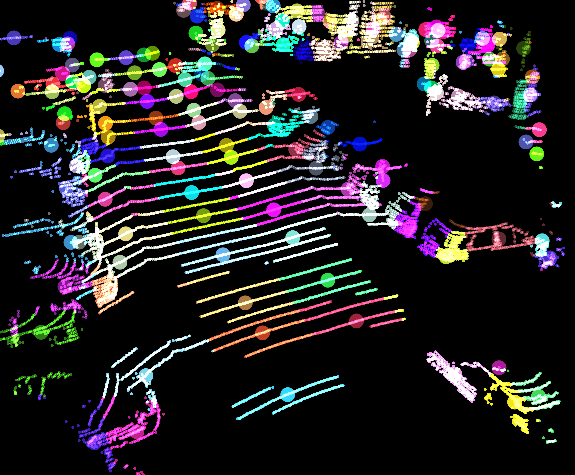}}
  \\
  \subfloat[Projection of the main points on the {RGB} image.]{%
    \includegraphics[width=0.95\linewidth]{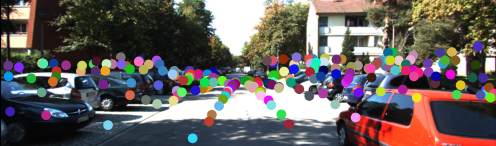}}
  \caption{Relationships between voxels on image frame and {BEV}
    frame. One voxel is represented with the same color on each
    figure. The {BEV} map is low resolution for better readability.}%
  \label{fig:voxels} 
\end{figure}

Aside from the feature vector, a ``Main Point'' is defined for each
voxel.  For $\chi = \{ \bm{x_{1}}, \ldots, \bm{x_{n}} \}$ the 3D
points inside the voxel, $\bm{\bar{x}}$ the mean point, $d$ the
Euclidean distance, the ``3D Main Point'' $\bm{x_{M}}$ is defined as
\begin{equation}
    \bm{x_{M}} = \arg \min_{\bm{y} \in \chi} d(\bm{y} - \bm{\bar{x}}).
\end{equation}
This point is representative of the content of the voxel while
belonging to the original set of points. The projections of the
estimated main points are computed so each voxel is
represented by one point on the image frame as illustrated in
Figure~\ref{fig:voxels}.

\subsection{Proposed Network Architecture}
\begin{figure*}[htb]    
  \centering%
  \subfloat{\includegraphics[width=0.95\linewidth]{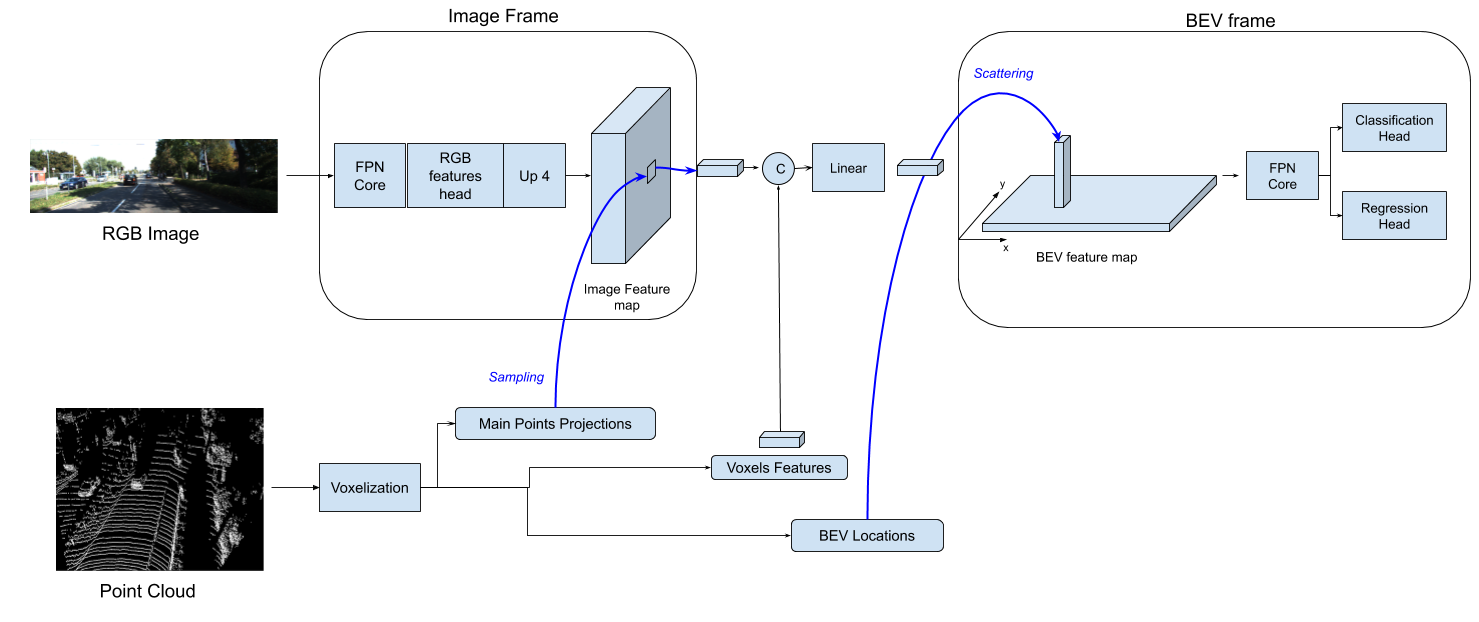}}
  \caption{Overview of the proposed network architecture.}%
  \label{fig:rpn_design} 
\end{figure*}

The architecture of the {RPN} can be broken down into two sub-parts:
an image feature extractor operating on the image frame and a {BEV}
estimator running on the {BEV} frame. Each block is
described in the following paragraphs. The global structure of the
{RPN} is described in Figure~\ref{fig:rpn_design}.

\subsubsection{Image Feature Extractor}
This network produces a feature map from the input {RGB} image. A
Feature Pyramid Network ({FPN}) architecture has been adopted to
extract information from feature maps at multiple scales. First, the
image is processed by a core network which returns three features maps
at different scales. These maps are given to a head network (named
``{RGB} feature head'' in Figure~\ref{fig:rpn_design}) whose task is
to merge these feature maps into one. The resulting map is up-sampled
by a factor 4 (the block ``Up4''). Using the projection of each
voxels' main point, the feature map is sampled and each voxel is
assigned with the extracted feature vector.

\subsubsection{{BEV} Estimator}
Each voxel has now a feature vector from the image feature encoder and
a feature vector from the 3D point cloud. Both are concatenated and
transformed by linear layers. Knowing their location on the {BEV}
grid, it can be filled with the resulting feature vectors.  A second
{FPN} core network is applied on the {BEV} feature grid. The resulting
feature maps are then given to two different head networks, one for
the classification, the other for the regression. The use of {FPN} is
motivated by the fact that objects from different categories may not
be represented with the same scale on both frames (close/far objects
in the image frame, small/large objects in the {BEV} frame). Feature
Pyramid Networks manage well objects with various apparent sizes as
multi-scale features maps are generated.

\subsection{Network Output}
Contrary to many methods defining prior boxes with various sizes and
orientation located in fixed positions, an anchor-free approach
inspired by~\cite{zhou2019objects} is employed here to represent the
objects where each predicted one is only described by its center.  In
fact, anchors-based methods require a set of hyper-parameters
specifically tuned to work correctly (size of anchors, {IoU} threshold
to define a positive anchor, etc.). Anchor-free method removes this
constraint, allowing to reduce the number of critical parameters that
may greatly affect the training process.

In common datasets, 3D obstacles are represented by a position
$(x, y, z)$, dimensions $(h, w, l)$ and an orientation $\theta$. The
orientation is restrained to the vertical axis for urban and
peri-urban scenes. However, our goal here is to extract regions with
defined limits and location but not the oriented boxes. At this stage,
estimating refined boxes is not suitable for two main reasons:
\begin{itemize}
\item {RGB} images provide mainly appearance features but estimating 3D
  spatial features from a projection is a difficult task. Advances in
  3D related tasks with monocular setups are still less precise than
  their counterparts using range sensors;
\item even if the pixels of an image are tightly packed, the sampling
  from the voxelization may not follow the resolution of the
  image. Some pixels are skipped and then the complete information is not
  obtainable.
\end{itemize}
In that sense, the regions of interest are defined as cuboids with a
specified side $s$ and the position of their center. The cuboid must
be large enough to encompass an important part of an object of
interest, but must not include other interfering objects.

The {RPN} returns two maps: A classification one and a regression one,
both having the same size. Each pixel of both map represents a cuboid
similar to a prior/anchor box, the first map gives its confidence
score, its category and an approximate location while the second map
gives the correction on the position and the predicted side. The
extraction of top proposals is summed up to the identification of
local maxima in the classification map and the gathering of the
estimated corrections on the regression map at the same locations.

\subsection{Training}
\subsubsection{Ground Truth Formatting}
Each ground truth object $i$ is defined by a semantic category and its
box parameters
$(x_{i}, y_{i}, z_{i}, h_{i}, w_{i}, l_{i} ,\theta_{i})$ with
$(x_{i}, y_{i}, z_{i})$ its position, $(h_{i}, w_{i}, l_{i})$ its
dimensions and $\theta_{i}$ its orientation.  Our {RPN} has to
estimate coarse {AABB} proposals that surround these ground truth
objects.

Ground truth maps are formatted using a process inspired
by~\cite{zhou2019objects}. Each label is represented by a Gaussian
function whose mean is the center of the object. A voxel is called
positive if its ground truth value is $1$ (the center of the object
belongs to this voxel). One label is then assigned to one voxel in the
classification map.

Concerning the regression map, let us denote $\Delta x$ the first
channel, $\Delta y$ the second one and $\Delta s$ the third one
representing the sides of proposals. Voxels are defined according to a
static grid with fixed position and each voxel on the classification
map represents a portion on the metric space. For the ground truth
object $i$, its associated voxel, whose center is located at
$(\tilde{x}_{i}, \tilde{y}_{i})$ in the metric space and at $(u, v)$
on the regression map, the corresponding target for the regression is
defined as:
\begin{align}
  \label{eqn:targets}
  \begin{split}
    \Delta x (u, v) &= \tilde{x}_{i} - x_{i}, \\
    \Delta y (u, v) &= \tilde{y}_{i} - y_{i}, \\
    \Delta s (u, v) &= \sqrt{{(w_{i})}^2 + {(l_{i})}^2}.
  \end{split}
\end{align}

\subsubsection{Loss Functions}
The {RPN} loss is a sum of two losses:
\begin{equation}
  L = \gamma_{cls}L^{cls} + \gamma_{reg}L^{reg}
\end{equation}
with $L^{cls}$, $L^{reg}$ the losses related to classification and
parameter regression, $\gamma_{cls}$, $\gamma_{reg}$ their respective
weights.

$L^{cls}$ being a pixel oriented focal loss on all elements (voxels)
for the scene level:
\begin{equation}
  L^{cls} = -\frac{1}{N} \sum_{uvc} \left\{
    \begin{array}{ll}
      {(1 - \hat{p}_{uvc})}^{\alpha} \log(\hat{p}_{uvc})  & if p_{uvc} = 1 \\
      {(1 - p_{uvc})}^{\beta} {(\hat{p}_{uvc})}^{\alpha} \log(1 - \hat{p}_{uvc})  & else
    \end{array}
  \right.
\end{equation}
with $\alpha, \beta$ being the hyper parameters that control the influence
of positive and negative voxels on the loss. $p_{uvc}, \hat{p}_{uvc}$
are respectively the ground truth and the prediction of the voxel
located in $(u,v)$ in the classification map for the class $c$. $N$
is the number of positive voxels in the ground truth map
  ($p_{uvc} = 1$).

$L^{reg}$ is defined as follows:
\begin{equation}
\begin{array}{lll}
 L^{reg} = -\frac{1}{N} \sum_{u_{pos}v_{pos}} 
 &   & SL1(\Delta x_{u_{pos}\ v_{pos}}, \hat{\Delta x}_{u_{pos}v_{pos}}) \\
 & + & SL1(\Delta y_{u_{pos}\ v_{pos}}, \hat{\Delta y}_{u_{pos}v_{pos}}) \\
 & + & SL1(\Delta s_{u_{pos}\ v_{pos}}, \hat{\Delta s}_{u_{pos}v_{pos}}) 
 \end{array}
\end{equation}
with $SL1(.,.)$ denoting the Smooth $L1$ loss, $(u_{pos}, u_{pos})$
the locations of the positive voxels in the regression maps,
$(\Delta x, \Delta y, \Delta s)$ the target maps defined earlier,
$(\Delta \hat{x}, \Delta \hat{y}, \Delta \hat{s})$ the predicted maps.

\subsection{Data Augmentation}%
\label{sec:data_augmentation}
This section describes the data augmentation strategies proposed for
the training. In autonomous driving field, the most common {LiDAR}
sensors used are rotating mechanical sensors, delivering 3D points
from one emission location in a layered manner. Points from the same
layer share the same latitude in a spherical frame centered on the
sensor. The different layers of the point cloud are supposed to be
registered. Then for each sample, some layers are randomly discarded
until a specific percentage of the point cloud layers remains.

Generally, images of open scenes are acquired with a large field of
view and high resolution as well. However, training with such
resolutions is cumbersome due to memory consumption aspects. Indeed,
during the training stage, gradients have to be computed on each pixel
at each layer of the neural network. GPUs have to store images and all
tensors computed by the network from them. In order to reduce these
phenomena, a random cropping strategy which is mostly used for
segmentation methods is adopted. Here, we set
$(\patchWidth, \patchHeight)$ for the width and the
height of the cropped region, respectively. The {RGB} encoder being fully
convolutional, feeding it on test time with an input with larger
resolution is possible as long as the image stays in the same domain
(lighting conditions, camera noise, etc.).

\begin{figure}[htb]
  \centering
  \subfloat[Selected label.\label{fig:data_aug_process_gt}]{%
    \includegraphics[width=0.90\linewidth]{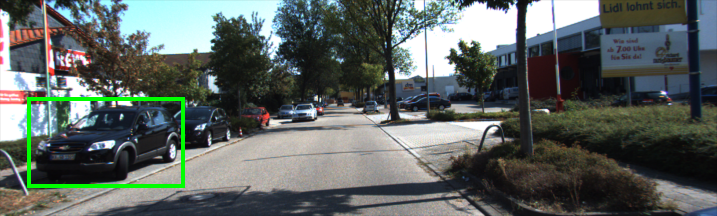}}
  \\
  \subfloat[Crop region centered.\label{fig:data_aug_process_center}]{%
    \includegraphics[width=0.90\linewidth]{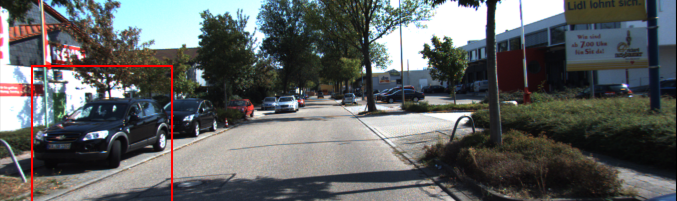}}
  \\
  \subfloat[Crop region shifted.\label{fig:data_aug_process_shift}]{%
    \includegraphics[width=0.90\linewidth]{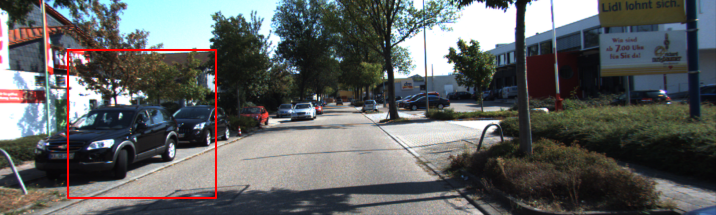}}
  \\
  \subfloat[Original {RGB} Image\label{fig:data_aug_rgb_original}]{%
    \includegraphics[width=0.60\linewidth,height=2cm]{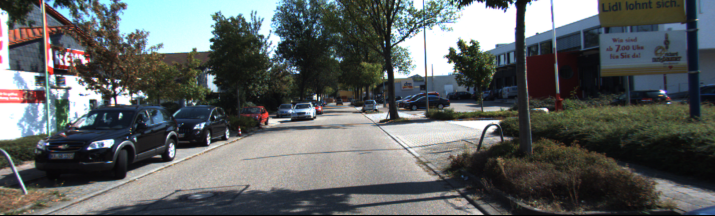}}
  \hfill
  \subfloat[Input {RGB} Image\label{fig:data_aug_rgb_final}]{%
    \includegraphics[width=0.32\linewidth,height=2cm]{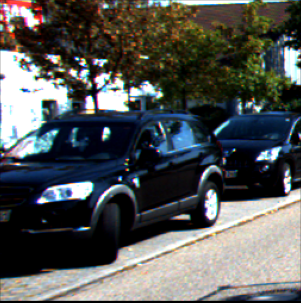}}
  \\
  \subfloat[Original point cloud (white) and Input point cloud (red)\label{fig:data_aug_pc}]{%
    \includegraphics[width=0.90\linewidth]{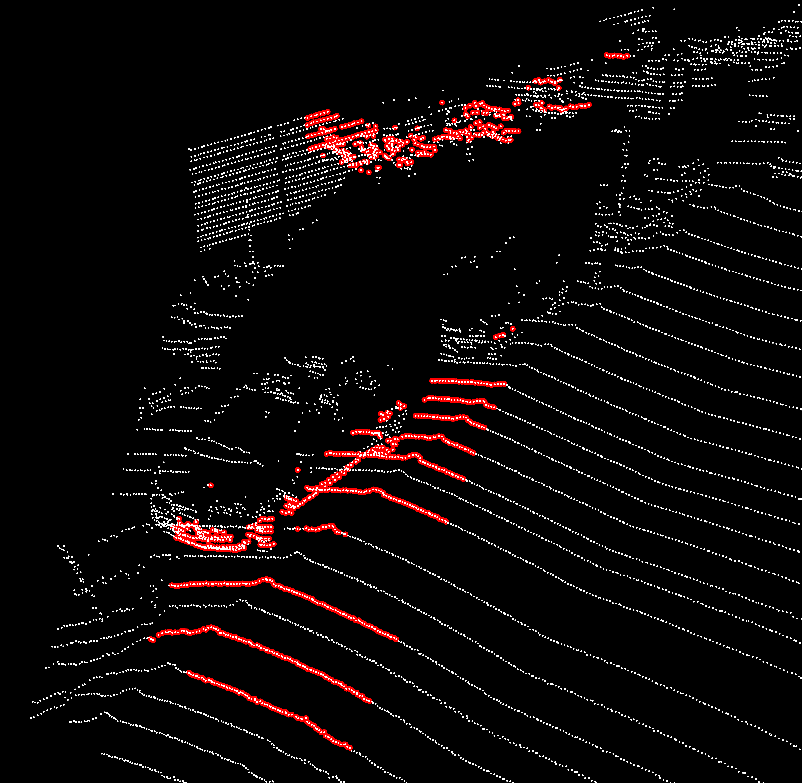}}
  \caption{Example of {KITTI} inputs for the training.}%
  \label{fig:data_aug} 
\end{figure}

Some categories of labels belong to the minority of labels or are
missing in most of the training samples. Purely random cropping may
lead to a majority of background or will mainly include the most
common category. Thus, the cropped region is chosen according to the
content of the scene:
\begin{itemize}
\item%
  if the scene contains at least one label, one of them can be randomly
  selected (Figure~\ref{fig:data_aug_process_gt}). A cropping region
  is then defined around the selected label
  (Figure~\ref{fig:data_aug_process_center}) and a random shift is
  applied so the selected label is not centered on the crop
  (Figure~\ref{fig:data_aug_process_shift});%
\item%
  if the sample does not contain labels, crops are selected randomly.
\end{itemize}

Even if the label is selected randomly, the probabilities between each
label are not uniform and depend on one or multiple criteria.
Random horizontal flips and color jittering are applied on the cropped
patch. 

To keep data consistency, every point whose projection is outside the
cropping region is discarded leading, for the point cloud and for the
{BEV} image, to a much smaller field of view.  Examples of the
resulting {RGB} and 3D input data are shown in
Figure~\ref{fig:data_aug_rgb_final} and Figure~\ref{fig:data_aug_pc},
respectively.

\section{Experiments and Results}%
\label{sec:experiments}
In this section, we first introduce the two datasets used in this
study, {KITTI} and {nuScenes}. The next section details the chosen
parameters for the proposed network and the training
stage. Thereafter, the conducted experiments and their results
are described. Finally, some ablation studies are
presented to justify the choices unrelated to the considered datasets.

\subsection{Datasets}%
\label{sec:experiments-datasets}

\subsubsection{KITTI}%
{KITTI}~\cite{Geiger2013vision} is a public dataset for autonomous
driving tasks available since 2012 that contains data recorded by an
equipped vehicle around Karlsruhe, in Germany. The embedded sensors
include a 64-layer {LiDAR} on the top of the vehicle and two front
color cameras. In this study, only the subset dedicated to 3D object
detection is used. This dataset contains \num{7481} annotated training
samples and \num{7518} testing samples, each sample
consists of a panoramic 3D point cloud, the {RGB}
images from each camera, the calibration parameters, all being
synchronized. The annotations are only defined in the camera's
field. As labels in the testing samples are not publicly available,
the original samples are split into \num{3741} training samples and
\num{3740} validation samples.

\subsubsection{nuScenes}%
{nuScenes}~\cite{nuscenes2019} is another public dataset first
released in 2019, mainly for object detection. Samples were collected
in Boston and Singapore. The acquisition vehicle boards $32$-layer
{LiDAR} sensor, $6$ color cameras, each one facing a different
direction, $5$ radars, a {GPS} and an {IMU}.  The dataset exists in
two versions:
\begin{itemize}
\item a complete version: it contains \num{1000} scenes split into
  \num{850} training scenes and \num{150} testing scenes;
    \item a mini version: it contains \num{10} scenes.
\end{itemize}
Each scene is \SI{20}{\second} long.  Samples are annotated at
\SI{2}{\hertz}. Intermediate point clouds called ``Sweeps'' are
defined between annotated samples.  The dataset defines 23 different
categories of objects.

\subsubsection{Experimental Setup}
All the presented networks are trained on the
{KITTI} train set. The following results are computed either on the
validation set or on the scenes of {nuScenes} Mini to keep the same
order of magnitude on the number of samples.  A common technique used
for {nuScenes} dataset is, for one sample, to accumulate some previous
sweeps, not labeled, and augment each point of the accumulated cloud
with its original timestamp, in order to get a denser point cloud. No
sweeps are available in the {KITTI} dataset, thus only the {nuScenes}
annotated point clouds are used.  Moreover, all the
{KITTI} scenes were captured on daylight. Hence, we focus on the
{nuScenes} sequences that were recorded on similar conditions and skip
night sequences as numerous new disturbances occur on the image (lens
flares due to streetlights and car lamps, higher noise, etc.).
Finally, if {KITTI} object detection benchmarks focus on three main
classes, {nuScenes} meanwhile defines more categories. Hence, in order
to allow the comparison between results on both datasets, we selected
the {nuScenes} categories that are the closest to the {KITTI}
classes. Table~\ref{table:kitti_nuscenes} illustrates the established
correspondences between classes from both datasets.

\begin{table}[H]
  \caption{Established correspondences between {KITTI} classes and
    {nuScenes} categories.}%
  \begin{center}
    \begin{tabular}{c | c}
      \toprule
      {KITTI} & {nuScenes}\\ 
      \midrule
      \mbox{Car} (and \mbox{Van}) & \mbox{Car} \\
      \mbox{Pedestrian} & \mbox{Pedestrian} \\
      \mbox{Cyclist} & \mbox{Bicycle}  attribute ``with rider'' \\
      \bottomrule
    \end{tabular}%
    \label{table:kitti_nuscenes}
  \end{center}
\end{table}

Concerning the {nuScenes} dataset, we focus the daylight scenes from
the version ``Mini''. This subset counted \num{4539} \mbox{Car}
labels, \num{2826} \mbox{Pedestrian} labels and \num{75}
\mbox{Cyclist} labels.

\subsection{Implementation Details}%
\label{sec:experiments-implementation}
In this section, details on the network parameters and on the training
procedure are provided.

\subsubsection{Details on the Network Architecture}
\begin{figure}[H]    
  \centering%
  \subfloat[FPN Core. Blocks named $Res_{i}$ are Resnet Blocks. Ratios
  on the right indicate the size of the feature map at each
  stage.\label{fig:fpn_core}]{%
    \includegraphics[width=0.95\linewidth,height=3cm]{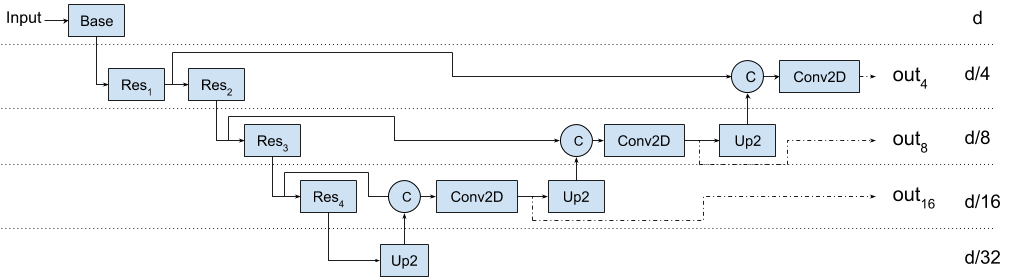}}%
  \\
  \subfloat[FPN Head.\label{fig:fpn_head}]{%
    \includegraphics[width=0.95\linewidth,height=3cm]{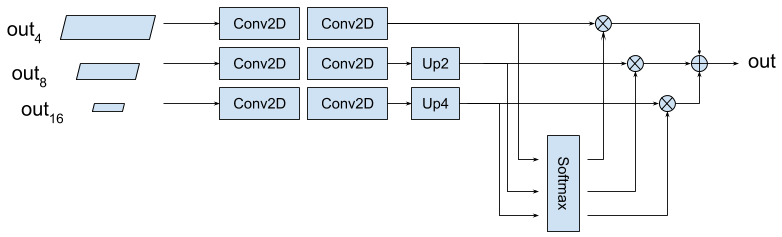}}%
  \caption{Details of {RPN} sub-networks.}%
  \label{fig:rpn_details} 
\end{figure}

As illustrated in Figure~\ref{fig:rpn_design}, the proposed network is
composed of two main blocks, an image feature extractor and a {BEV}
estimator. Each of them is composed of instances of the two
sub-networks, a {FPN} core (Figure~\ref{fig:fpn_core}) whose role is
to extract three feature maps at different scales and a {FPN} head
(Figure~\ref{fig:fpn_head}), in charge of merging those features maps
into one output. Blocks labeled as ``{RGB} feature head'',
``Classification head'' and ``Regression head'' are all instances of a
{FPN} head. Their structures are detailed in
Figure~\ref{fig:rpn_details}.

\mbox{Conv2D{($c_{in}$, $c_{out}$, $k$, $s$, $p$)}} refers to a 2D
convolution operator with $c_{in}$, $c_{out}$ the number of input and
output channels, $k$ the kernel size, $s$ the stride, $p$ the
padding. Likewise, \mbox{MaxPool{($k$, $s$, $p$)}} denotes a Max Pooling
operator, $k$ the kernel size, $s$ the stride, $p$ the padding. Blocks
named ``Up2'' and ``Up4'' represent an up-sampling with a factor 2 and
4, respectively.

\textbf{Image Feature Extractor}\\
The block named ``Base'' used in the Image Feature Extractor's Core is
composed of a \mbox{Conv2D{(3, 64, 7, 2, 3)}}, a batch normalization, a
\mbox{ReLU} activation and a \mbox{MaxPooling{(3,2,1)}}.  Block labels
$Res_{i}$ comes from a \mbox{ResNet18} architecture.  On the {FPN}
core (Figure~\ref{fig:fpn_core}), $out_{16}$, $out_{8}$, $out_{4}$ are
produced by \mbox{Conv2D{(768, 256, 1, 1, 0)}}, \mbox{Conv2D{(384, 128,
  1, 1, 0)}} and \mbox{Conv2D{(192, 64, 1, 1, 0)}}, respectively.
Concerning the {RGB} feature head as illustrated in
Figure~\ref{fig:fpn_head}, each feature map are given to their own
\mbox{Conv2D{($c_{dim}$, 64, 3, 1, 1)}} followed by a \mbox{Conv2D{(64,
  $c_{task}$, 1, 1, 0)}}, where $c_{dim}$ depends on the numbers of
channels of the feature map and $c_{task}$. For $out_{16}$, $out_{8}$,
$out_{4}$, $c_{dim}$ is set to $256$, $128$ and $64$,
respectively. $c_{task}$ is set to $64$.

\textbf{{BEV} Estimator}\\
The proposed architecture is close to the one described in the previous
paragraph entitled ``Image Feature Extractor''. As a result, only the
differences between Image Features Extractor architecture and the
{BEV} Estimator structure are described.  The first
convolution in the ``Base'' block is replaced by a \mbox{Conv2D{(64,
    64, 7, 2, 3)}}.  Classification and Regression heads have the same
structure as the {RGB} feature head, only $c_{task}$ varies. For the
classification, $c_{task} = 3$ is selected for the studied classes:
\mbox{Car}, \mbox{Pedestrian} and \mbox{Cyclist}. For the regression,
$c_{task}$ is assigned to $3$ for estimating $(x, y, s)$.

Table~\ref{table:global_params} defines the borders of the search
space and the dimensions of the input maps (from the voxelization) and
output maps (the different head sub-networks).
\begin{table}[H]
\caption{Global parameters (train and test).}%
  \begin{center}
    \begin{tabular}{c c}
      \toprule
      Parameters & Values \\ 
      \midrule
      ($\sceneXMin$, $\sceneXMax$) & (\SI{0}{\meter}, \SI{50}{\meter})  \\
      ($\sceneYMin$, $\sceneYMax$) & (\SI{-25}{\meter}, \SI{25}{\meter}) \\
      ($\sceneXRes$, $\sceneYRes$) & (\SI{0.0625}{\meter}, \SI{0.0625}{\meter}) \\
      ($\sceneXNum$, $\sceneYNum$) & (800, 800)\\
      ($\sceneXResOut$, $\sceneYResOut$) & (\SI{0.25}{\meter}, \SI{0.25}{\meter}) \\
      ($\sceneXNumOut$, $\sceneYNumOut$) & (200, 200)\\
      \bottomrule
    \end{tabular}%
    \label{table:global_params}
  \end{center}
\end{table}

The values displayed in Table~\ref{table:global_params}, are selected
according to the labels existing on the {KITTI} and the {nuScenes}
datasets. All data are annotated between \num{0} and \SI{50}{\meter}
around the sensor, except for cars in {KITTI} which are annotated
until \SI{80}{\meter}.  \mbox{Pedestrian}s are the smallest targets
studied in this work, the output voxel size is selected, so they can
cover at least one voxel on the output maps.

\subsubsection{Training Stage Parameters}
The focus criteria for the patch extraction in the data augmentation
is the 3D volume: The smaller the target is, the greater the
probabilities to be selected are. As small objects are harder to
detect, a special focus allows to improve the attention of the network
to details. Dimensions of the {RGB} patches
$(\patchWidth, \patchHeight)$ are $(256, 256)$. Between $20\%$ and
$40\%$ of the layers from point clouds are kept.
$(\gamma_{cls}, \gamma_{reg})$ are defined as $(1, 1)$. For the focal
loss, $(\alpha, \beta)$ are defined as $(2, 4)$. These values come
from an improved implementation of~\cite{yan2018second}.  All models
are trained for $100$ epochs using an Adam optimizer and a one-cycle
learning rate scheduler with a maximum learning rate set to $0.001$.

\subsection{Experiments Conditions Description}%
\label{sec:experiments-network}
\subsubsection{Three Configurations of the Proposed Network}%
In order to highlight the relevance of the proposed network, its
performances are presented and discussed under
multiple configurations and training setups. For that purpose, three
configurations of {R-AGNO-RPN} are trained and studied according to
the considered sensors:
\begin{itemize}
\item an image with 3D Point cloud {RPN}, its architecture is detailed in
  Figure~\ref{fig:rpn_design}, the prefix $Cam3D\_$ is employed to
 identify it;
\item an image {RPN}, which has the same architecture as the previous
  one but the voxel features are not retained. However, the point cloud
  information required for the sampling and the scattering operations
  are still used. This experiment is named with the prefix $Cam\_$ in
  the following sections;
\item a {LiDAR}-based {RPN}, the image frame part is removed, only the
  point cloud features and the voxels locations are used, named with
  the prefix $3D\_$ in the following sections.
\end{itemize}
These three configurations are trained according to two
following procedures:
\begin{itemize}
\item the network is trained from all the point cloud layers,
  identified with the suffix $\_all$; 
\item the network is trained from randomly selected point cloud
  layers, between $20\%$ and $40\%$ of the layers are kept, identified
  with the suffix $\_few$.
\end{itemize}


For the inference, only the \num{20} top proposals are selected as
the greatest part of scenes contain at most \num{15}
labels.

\subsubsection{Comparison with a State-of-the-Art Network}%
Still with the aim of highlighting the strength of the proposed
algorithm, we also propose to carry out a comparative study with a well-known reference
state-of-the-art method, the {PointPillars}
network~\cite{Lang_2019_CVPR} which has popularized the use of column voxels. However, in order to conduct the study under comparable
conditions, some modifications were implemented:
\begin{itemize}
\item the three classes are simultaneously managed. In the original
  paper~\cite{Lang_2019_CVPR}, a set of weights was especially trained
  for \mbox{Car}s and another one was optimized to manage
  \mbox{Pedestrian}s and \mbox{Cyclist}s;
\item to match with our outputs, only axis-aligned boxes are
  returned. By removing the estimation of some parameters such as
  the orientation during the training, the loss and the network
  focus mainly on the localization;
\item the anchors are modified to become cuboids whose side is the
  diagonal of the original prior boxes.
\end{itemize}
    
The {PointPillars} network is also trained with original data and
altered data. These two experiments are labeled $PP\_all$ and
$PP\_few$, respectively. Everything else stayed the same as in the
original paper (network architecture, training process, etc.).  The
term ``Number Layers'' refers to the number of layers in the point
cloud used at test time.

\subsection{Experimental Results Analysis}%
\label{sec:experiements-evaluation}
We evaluate in this section our proposed networks through Average
Precision (\mbox{AP}) on the validation set. \mbox{AP}
takes into account the false positives and prediction scores to
reflect how close the results are from the labels. Even if the
developed network aims to be the first stage of a two-stage detector, \mbox{AP} gives an indication on how confident the
network is.

\mbox{AP} is computed on {AABB} (axis-aligned bounding
boxes). \mbox{AP} is computed for all classes with an
{IoU} threshold of $0.5$.

\begin{figure*}[t]
  \centering
  \subfloat[\mbox{AP Car}\label{fig:ap_comparisons_car}]{%
    \includegraphics[width=0.49\linewidth]{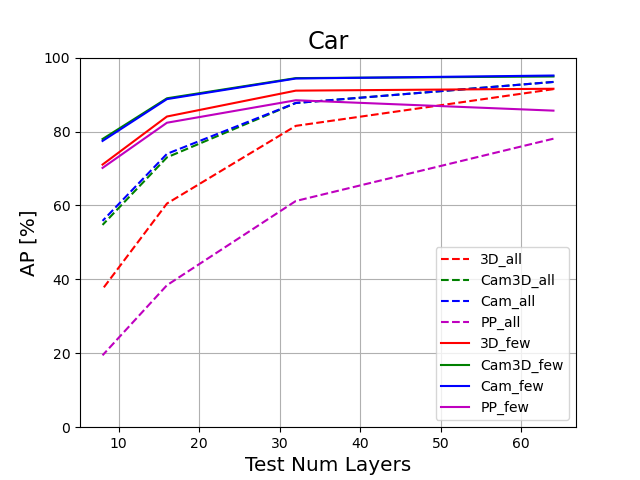}}
  \hfill%
  \subfloat[\mbox{AP Cyclist}\label{fig:ap_comparisons_cyclist}]{%
    \includegraphics[width=0.49\linewidth]{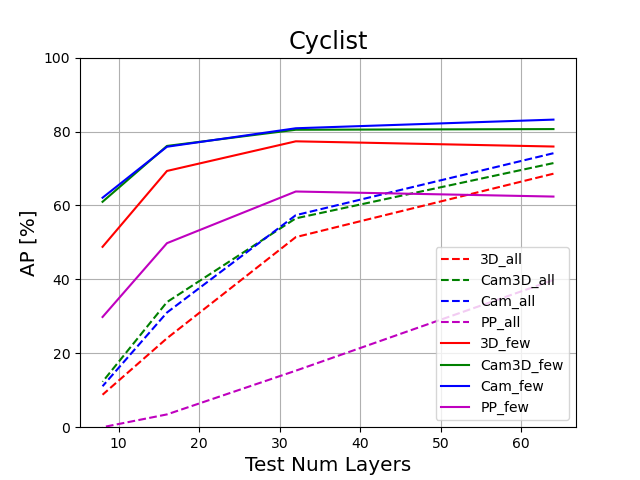}}
  \\
  \subfloat[\mbox{AP Pedestrian}\label{fig:ap_comparisons_pedestrian}]{%
    \includegraphics[width=0.49\linewidth]{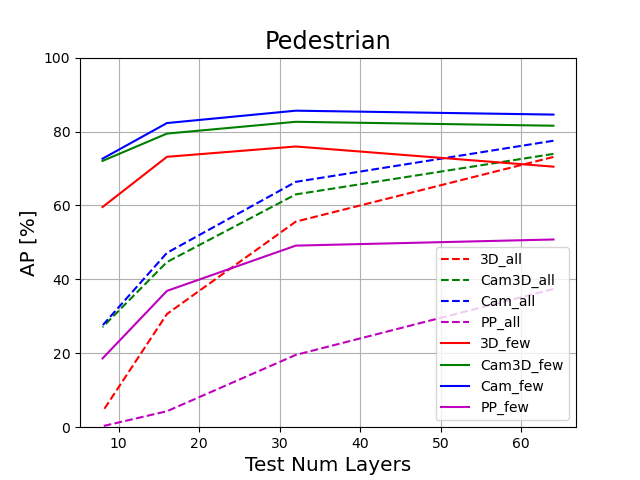}}
  \hfill
  \subfloat[\mbox{AP} mean\label{fig:ap_comparisons_mean}]{%
    \includegraphics[width=0.49\linewidth]{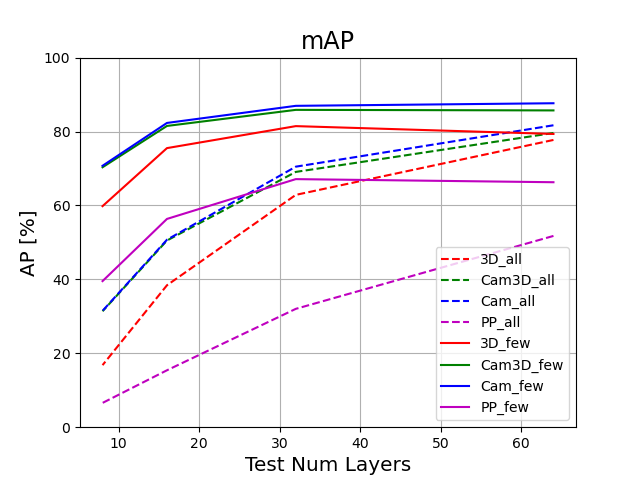}}
  \caption{Evolution of the Average Precision (IoU \num{0.5}) depending on the number of
    layers in point clouds.}%
  \label{fig:ap_comparisons} 
\end{figure*}

The results on average precision are illustrated in
Table~\ref{table:ap_kitti} and Figure~\ref{fig:ap_comparisons}
represents the evolution of \mbox{AP} for different trainings with the number
of layers of input point clouds on testing phase. The layers are
selected in a deterministic way so all the experiments can be realized
under the same conditions.

\begin{table}[!t]
\caption{Average Precision (IoU \num{0.5}) on {KITTI} \textit{Validation} set.}
  \begin{center}
    \begin{tabularx}{0.5\textwidth}{l | X | X X X | X}
      \toprule
      Input & Testing number layers & \mbox{Car} & \mbox{Pedestrian} & \mbox{Cyclist} & mAP\\ 
      \midrule
      $PP\_all$ &  \multirow{6}{*}{64} & $78.06$ & $37.39$ & $39.82$ & $51.75$\\
      $PP\_few$ &  & $85.65$ & $50.79$ & $62.40$ & $66.28$\\
      
      $3D\_all$ &  & $91.46$ & $73.12$ & $68.59$ & $77.72$\\
      $3D\_few$ &  & $91.57$ & $70.49$ & $75.94$ & $79.34$\\
      
      $Cam3D\_all$ &  & $93.42$ & $73.95$ & $71.46$ & $79.61$\\
      $Cam3D\_few$ &  & $94.89$ & $81.57$ & $80.66$ & $85.71$\\
      
      $Cam\_all$ &  & $93.41$ & $77.51$ & $74.13$ & $81.69$\\
      $Cam\_few$ &  & $\bm{95.15}$ & $\bm{84.58}$ & $\bm{83.22}$ & $\bm{87.65}$\\
    
      \midrule
      $PP\_all$ & \multirow{6}{*}{8} & $19.45$ & $0.27$ & $0.0$ & $6.57$\\
      $PP\_few$ &  & $70.15$ & $18.61$ & $29.82$ & $39.53$\\
      
      $3D\_all$ &  & $37.37$ & $4.23$ & $8.78$ & $16.79$\\
      $3D\_few$ &  & $71.06$ & $59.57$ & $48.8$ & $59.81$\\
      
      $Cam3D\_all$ & & $54.73$ & $27.08$ & $12.29$ & $31.37$\\
      $Cam3D\_few$ & & $\bm{77.97}$ & $72.05$ & $60.99$ & $70.33$\\
      
      $Cam\_all$ & & $55.82$ & $27.54$ & $11.08$ & $31.48$\\
      $Cam\_few$ & & $77.45$ & $\bm{72.65}$ & $\bm{62.09}$ & $\bm{70.73}$\\
      
      \bottomrule
    \end{tabularx}%
    \label{table:ap_kitti}
  \end{center}
\end{table}

\subsubsection{Contributions of the Training Procedure}
Figure~\ref{fig:ap_comparisons} illustrates two different types of
curve variations depending on the conditions of the training with
unchanged ($\_all$) or with modified data ($\_few$).  Labeled
experiments $\_few$ tend to be more resilient to the reduction of
input layers than their counterparts labeled $\_all$.  For instance,
for the {PointPillars} network ($PP\_$) on the ``\mbox{Car}'' class
(Figure~\ref{fig:ap_comparisons_car}), the difference between the
maximum and the minimum measured \mbox{AP} values is $59.06$ on $PP\_all$
but decreases to $15.5$ on $PP\_few$.  Moreover, it can be noticed
that even if $\_few$ experiments are trained on fewer data, they
perform better than their equivalent $\_all$ on 64 layers {LiDAR}. The
assumption is that for $\_few$ experiments, the network learns
features that summarize better discriminant aspects than $\_all$
experiments as it must identify the targets on fewer data. In that
sense, every new point adds information to an already compact but
explicit descriptor. These results demonstrate the contribution of our
data augmentation procedure on the robustness with respect to
variations on the number of layers.

\subsubsection{Comparison between 3D Point Cloud Models}
In this section, {PointPillars} and the $3D\_$ version of {R-AGNO-RPN}
are compared in order to evaluate the proposed architecture. The focus
is on experiments labeled \_$all$.  Whatever the number of layers, our
network provides better results than {PointPillars}. However, the
difference of performances is more obvious on small targets
(\mbox{Pedestrian}, Figure~\ref{fig:ap_comparisons_pedestrian} and
\mbox{Cyclist}, Figure~\ref{fig:ap_comparisons_cyclist}). The same
observations can be noticed on experiments labeled $\_few$.
PointPillars manages less but bigger voxels that are processed
independently, while $3D\_$ generates more but smaller voxels that are
directly linked to their respective neighbors through the first
convolution. Moreover, contrary to {PointPillars} that concatenates the
features maps from different scales before estimating outputs, the
{FPN} head sub-networks estimate outputs for each feature maps, and
then merge them into one output. As a result, each scale can be
specialized for a specific type of targets.  These results demonstrate
the relevancy of our architecture that performs better than
PointPillars on the same input data.

\subsubsection{Contribution of {RGB} Images}
Table~\ref{table:ap_kitti} shows that without data augmentation for
the training and using the 64 layers at test time, $Cam\_all$ and
$Cam3D\_all$ reach respectively \mbox{AP} of $81.69\%$ and $79.61\%$
while $3D\_all$'s \mbox{AP} is $77.72\%$. The difference of
performance is accentuated when fewer points are available ($31.48\%$,
$31.37\%$ and $16.79\%$, respectively). This drop in performance can
be explained as on purely point cloud methods, the result heavily
depends on the number of points that are available.  Images are 
regular structures so each pixel of a feature map is
highly correlated with its neighbors in the image frame. Even if the
point cloud is sparser, only the number of sampled pixels vary, so it
is still possible for the {RGB} feature encoder to tell the difference
between a pole and a pedestrian. When these image features are
scattered on the {BEV} map, they already encode information about what
could be around the voxel.  The input grid, defining the number of
voxels and then the number of sampling points, has a fine resolution
($800 \times 800$). In this way, each feature sampled from the {RGB}
map is relocated in space with an acceptable precision. The category
of the obstacle may provide a relevant prior for the dimensions of the
box.  $Cam3D\_all$ and $Cam3D\_few$ are slightly less efficient than
$Cam\_all$ and $Cam\_few$.  In $Cam\_$ experiments, point clouds only
intervene for the mapping between the image frame and the {BEV}
frame. However, in $Cam3D\_$ experiments the point cloud features are
used as inputs for the linear network before the projection on the
{BEV} frame. While point cloud features only encode 3D points located
inside each voxel, {RGB} features already encode high level
information that takes account of the neighborhood of the sampling
point. Direct fusion at the voxel level may then act as a disturbance
for the network because of the difference of information. A latter
fusion on intermediate features or a matching between neighborhoods
between frames may correct this disturbance.

This experiment shows that {RGB} images, through the type of
information suitable for target identification, can be efficient in
{BEV} object localization if their features can be correctly
re-positioned in 3D space. Moreover, even on sparse point clouds, the
evaluations employing {RGB} images
provide the best results.

\subsubsection{Qualitative Results}
\begin{figure*}[htb]
  \centering
  \subfloat{\includegraphics[width=0.24\linewidth]{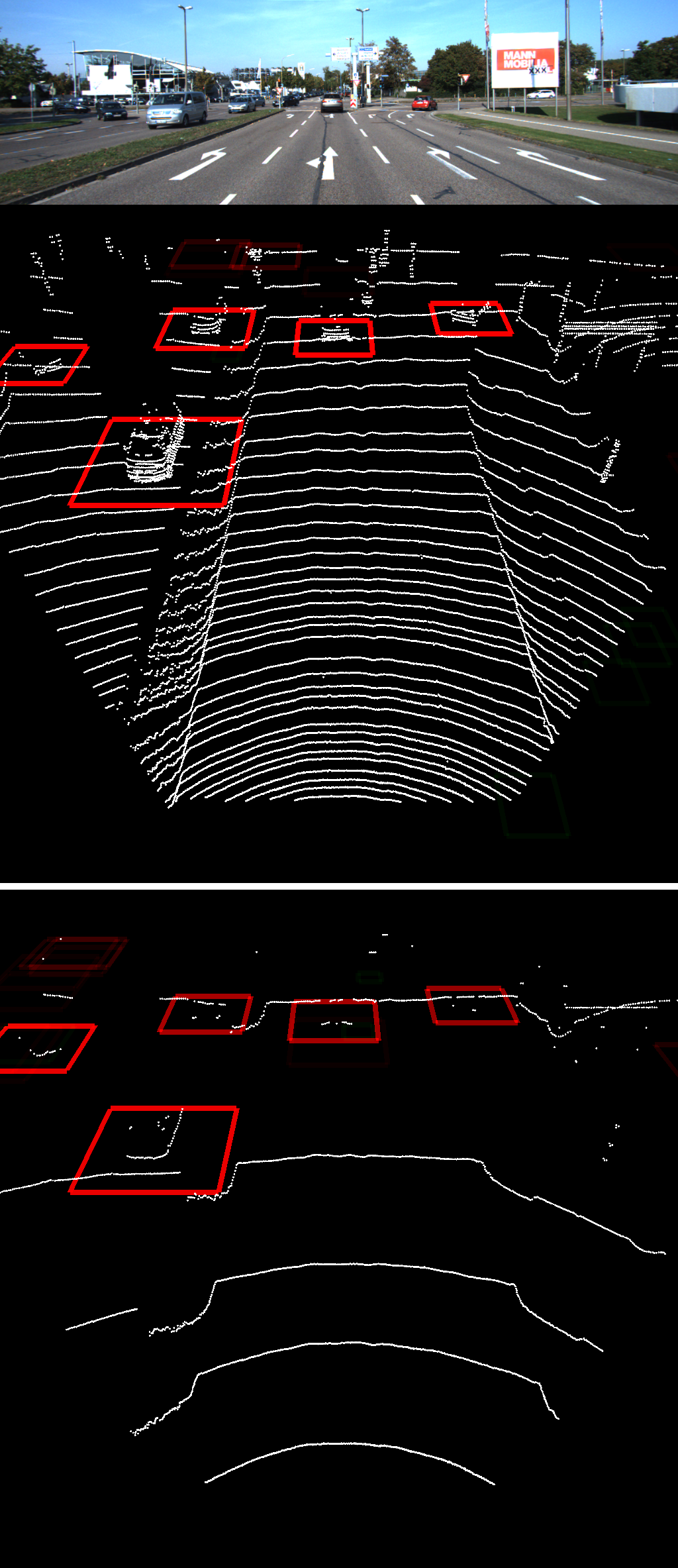}}
  \hfill%
  \subfloat{\includegraphics[width=0.24\linewidth]{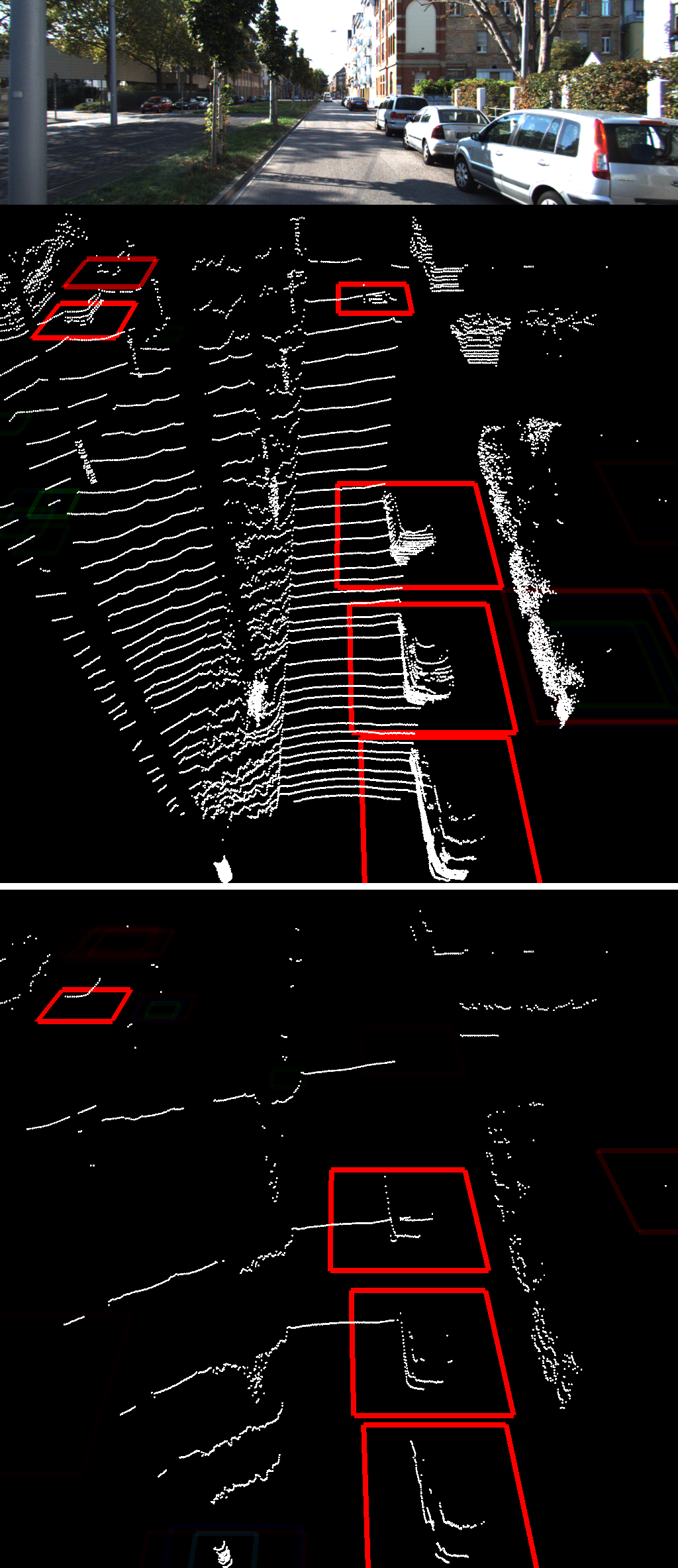}}
  \hfill
  \subfloat{\includegraphics[width=0.24\linewidth]{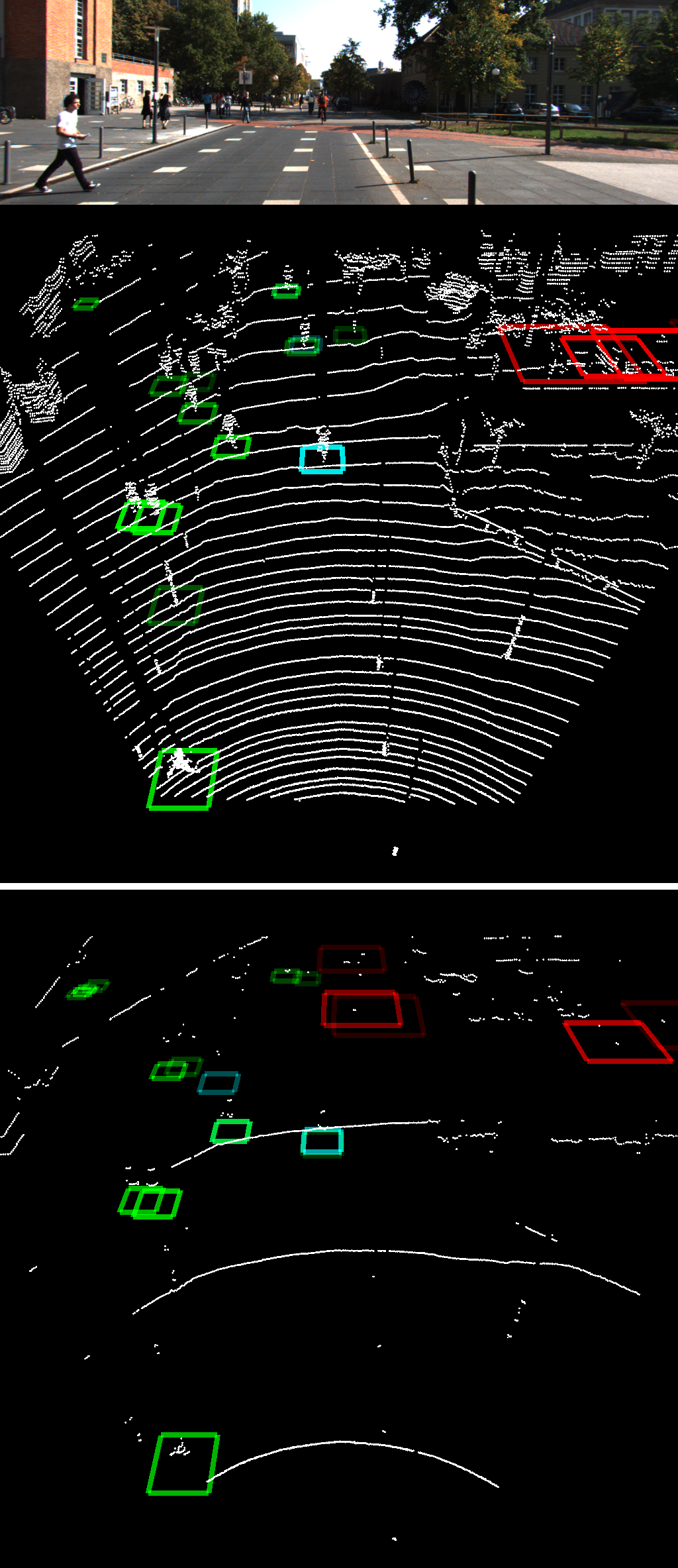}}
  \hfill
  \subfloat{\includegraphics[width=0.24\linewidth]{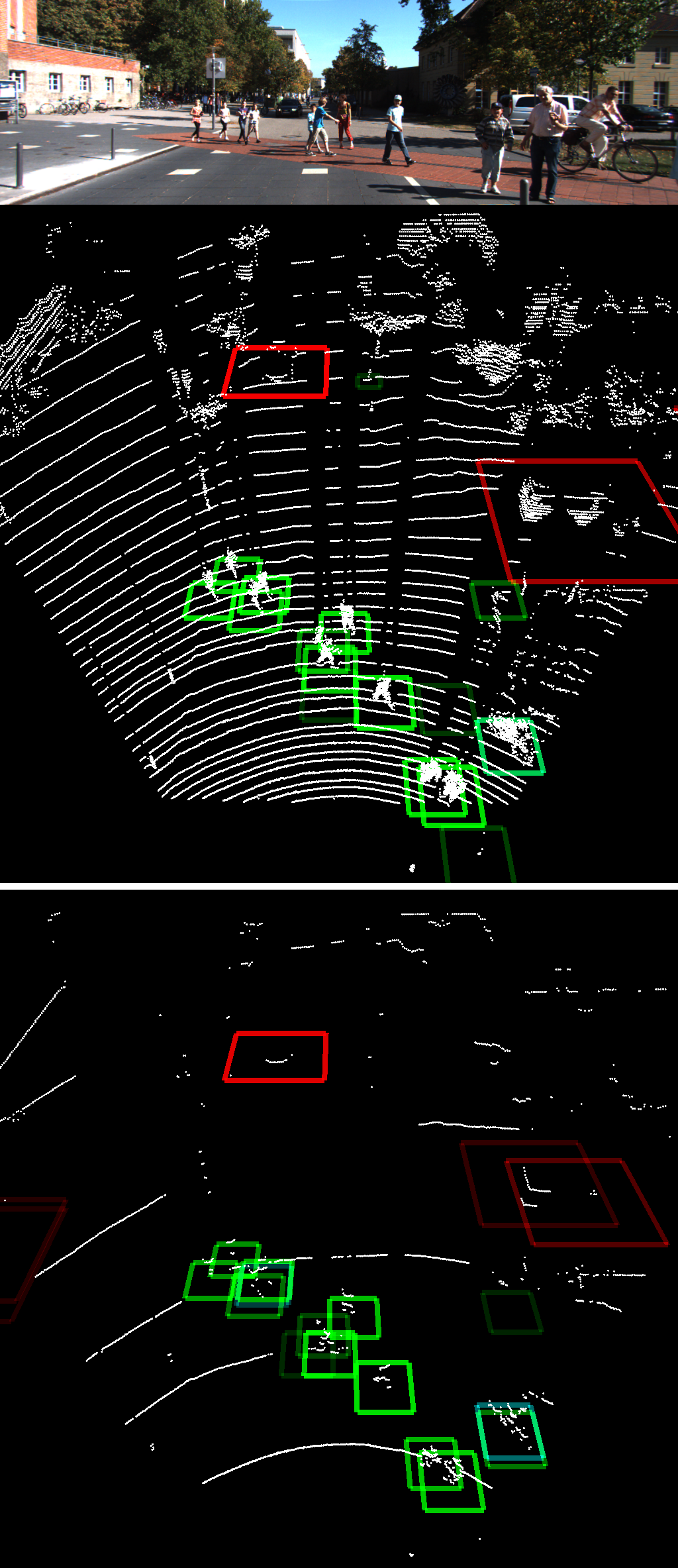}}
  \caption{Qualitative results on {KITTI} \textit{test} set. The first
    row contains input images, the second row contains original input
    point cloud, the third row contains examples with consequent
    reduction on input point clouds. Colors and transparency of boxes
    represent the estimated class and the confidence score (red: \mbox{Car},
    green: \mbox{Pedestrian}, cyan: \mbox{Cyclist}).}%
  \label{fig:kitti_qualitative} 
\end{figure*}

Figure~\ref{fig:kitti_qualitative} illustrates some results provided
by $Cam\_few$ on an unaltered point cloud and a highly trimmed point
cloud from \textit{test} split. Despite the important information loss
on the input point cloud, the proposed network is still able to
deliver correct proposals with high confidence, even if the scores are
generally smaller on reduced data.  Due to the reduction of 3D
points when layers are removed, some obstacles are no more
hit by laser impacts and then are no more detected by
the network as no sampling point reaches their representation in the
{RGB} image.  In some cases, our network produces more false positives
on incomplete data. As the image features are not affected by data
reduction, the assumption is that the {BEV} part still focuses more on
the arrangement of the voxels instead of the sampled features. In that
sense, a disturbance in the global arrangement may produce unexpected
outputs.

\subsection{Ablation Study}

While previous sections focused on the contributions of the {RGB} image
and the training procedure, the following paragraphs introduce the
results of the ablation study.

\subsubsection{Effects of the Merging Method in {FPN} Head}
In this paragraph, the impact of the merging method in the {FPN} head
is evaluated. Only ($Cam\_few$) is discussed here. Results are
detailed in Table~\ref{table:fpn_merge_op}.

\begin{table}[H]
  \caption{Comparison between $\Max$ and \textit{Softmax} operators for feature maps merging in {FPN} heads.}
  \begin{center}
    \begin{tabular}{c | c | c  c  c | c}
      \toprule
      Method & Test Number layers & \mbox{Car} & \mbox{Pedestrian} & \mbox{Cyclist} & mAP \\ 
      \midrule
      \textit{Max} & \multirow{2}{*}{64} & $\bm{95.26}$ & $34.11$ & $32.2$ & $53.86$ \\
      \textit{Softmax} &  & $94.78$ & $\bm{83.59}$ & $\bm{81.86}$ & $\bm{86.74}$ \\
      \midrule
      \textit{Max} & \multirow{2}{*}{8} & $\bm{77.46}$ & $31.00$ & $25.1$ & $44.52$ \\
      \textit{Softmax} &  & $77.00$ & $\bm{71.97}$ & $\bm{60.98}$ & $\bm{69.98}$ \\
      \bottomrule
    \end{tabular}%
    \label{table:fpn_merge_op}
  \end{center}
\end{table}

On both test conditions, the network operating with a $\Max$ operator
performs slightly better on \mbox{Car}s. However, its performances
drop on \mbox{Pedestrian}s and \mbox{Cyclist}s. As \mbox{Car}s are the
largest objects among the studied targets, they
  tend to be easier to detect even on low spatial resolution feature
  maps (such as $out_{16}$). Targets labeled with the categories \mbox{Pedestrian}s and \mbox{Cyclist}s
  will be more visible on more defined features maps. Features
  dedicated to small targets are then overwhelmed when using a $\Max$
  operator. The \textit{Softmax} operator helps to weight the features
  so smaller objects can still have an impact in the merging process.

\subsubsection{Impact of the Input Trimming Patterns}
The effects of different patterns are evaluated for the point cloud
sub-sampling.  For this experiment, the same network ($Cam\_$) is
trained through three trimming patterns:
\begin{itemize}
\item random uniform sampling ($20\%-40\%$ of the points);
\item random layer sampling ($20\%-40\%$ of the layers).
\end{itemize}

\begin{figure}[htb]
  \centering
  \subfloat[Original point cloud.]{\includegraphics[width=0.95\linewidth]{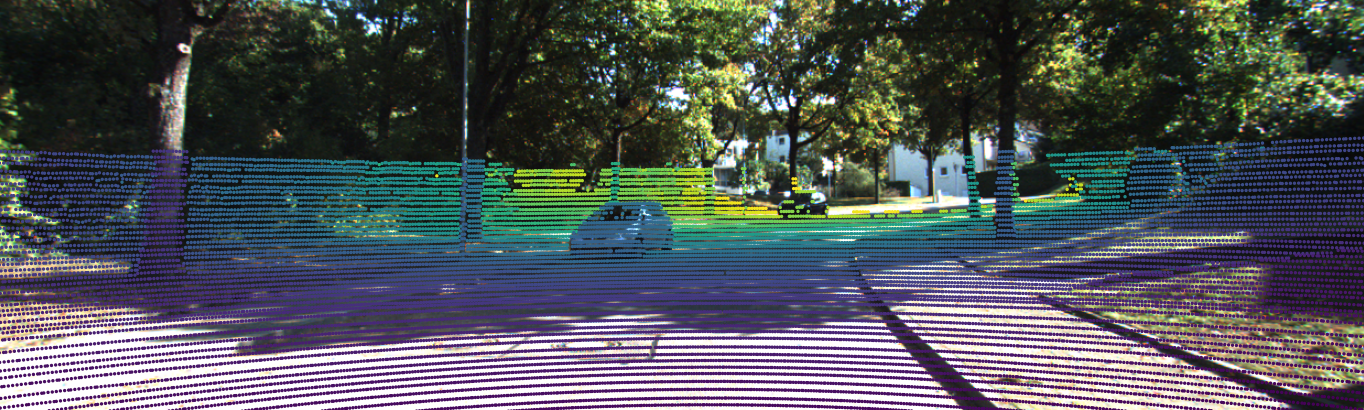}}
  \\%
  \subfloat[Uniform sub-sampling]{\includegraphics[width=0.95\linewidth]{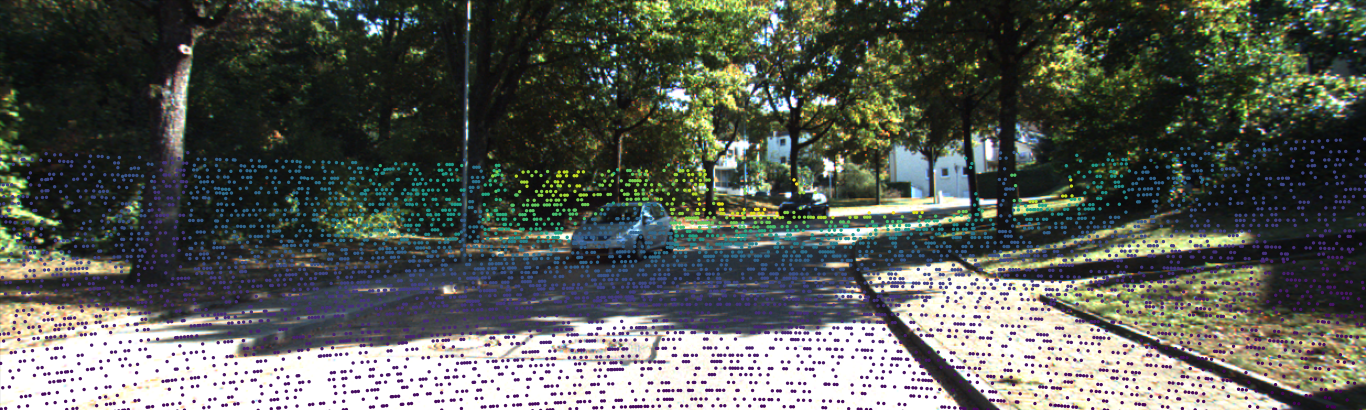}}
  \\
  \subfloat[Layer sub-sampling]{\includegraphics[width=0.95\linewidth]{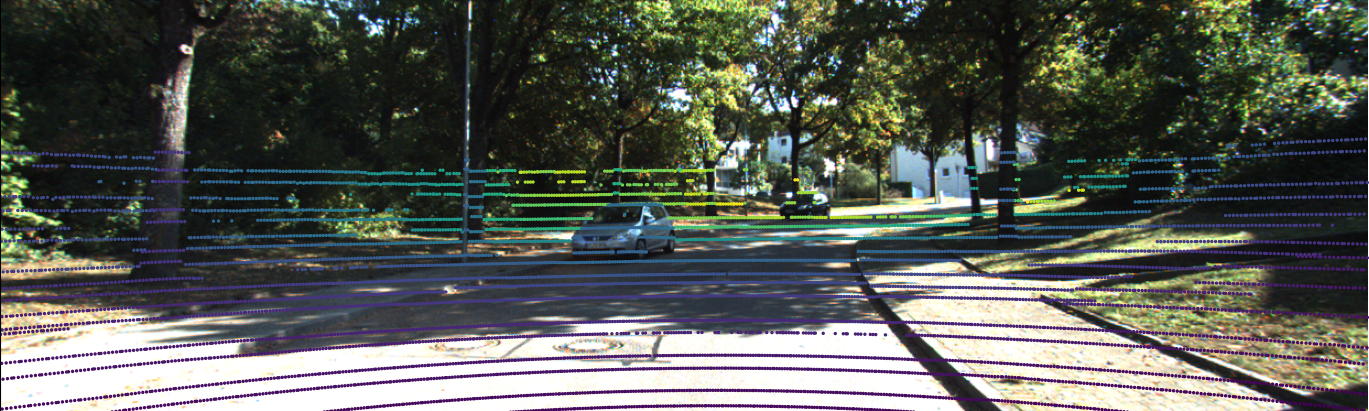}}
  \caption{Comparison between sampling patterns.}%
  \label{fig:samplings} 
\end{figure}

These two sampling methods are illustrated in
Figure~\ref{fig:samplings}. The results of the two experiments are
illustrated in Table~\ref{table:samplings}.

\begin{table}[H]
\caption{Influence of sampling patterns on the performance.}%
  \begin{center}
    \begin{tabular}{c | c | c  c  c}
      \toprule
      Class & Test Number layers & \mbox{Car} & \mbox{Pedestrian} & \mbox{Cyclist}\\ 
      \midrule
      Uniform & \multirow{2}{*}{64} & $95.12$  & $84.29$ & $80.74$ \\
      Layers & & $\bm{95.15}$ & $\bm{84.58}$ & $\bm{83.22}$ \\
      \midrule
      Uniform & \multirow{2}{*}{8} & $67.99$  & $50.20$ & $36.64$ \\
      Layers & & $\bm{77.45}$ & $\bm{72.65}$ & $\bm{62.09}$ \\

      \bottomrule
    \end{tabular}%
    \label{table:samplings}
  \end{center}
\end{table}

Uniform sub-sampling slightly degrades the performances on 64-layer
point clouds. Even if the uniform sub-sampling performs a reduction of
the number of points on the training stage, the layered structure of
the point cloud delivered by a mechanical {LiDAR} is no longer
respected. This causes the slight drop in performance on unaltered
data.  Somehow, the training domain and the testing domain do not
correspond when using uniform sub-sampling. This difference of domains
is highlighted when 8-layer point clouds are used.

\subsection{Transfer to {nuScenes} dataset}

\begin{figure}[!h]
  \centering
  \subfloat{\includegraphics[width=0.49\linewidth]{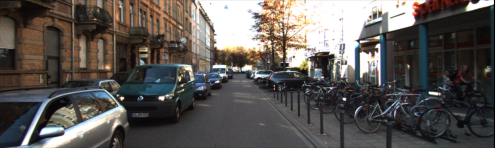}}
  \hfill%
  \subfloat{\includegraphics[width=0.49\linewidth]{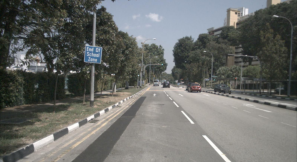}}
  \\
  \subfloat{\includegraphics[width=0.49\linewidth, valign=t]{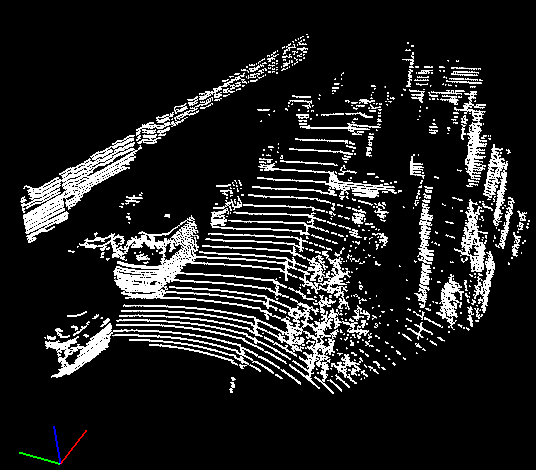}}
  \hfill
  \subfloat{\includegraphics[width=0.49\linewidth, valign=t]{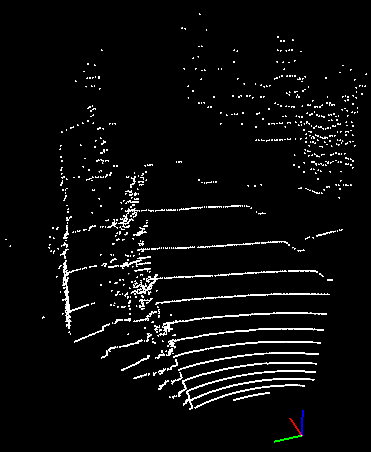}}
  \caption{Comparison between {KITTI} data (left) and {nuScenes} data (right).}%
  \label{fig:kitti_nusc_compare} 
\end{figure}

This section presents inference results on the {nuScenes} dataset
after pre-training on the {KITTI}'s one in order to highlight the
{R-AGNO-RPN}'s portability. The main difficulty in this experiment is
that two domains are changed, the image domain and the {LiDAR}
domain. A sample of each dataset is illustrated in
Figure~\ref{fig:kitti_nusc_compare}. Images in {nuScenes} are larger
and tend to be duller and are noisier. Moreover, even if data
represent urban scenes, the environments captured for {KITTI} greatly
differ from the ones captured for {nuScenes}. Concerning the point
cloud, even if the used {LiDAR} is a 32-layer sensor, the latitude
range is greater than the 64-layer {LiDAR} used in {KITTI}. As a
result, fewer layers (and then 3D points) are reserved to the ground
or objects at eye level.

In order to ensure similar experiment conditions
in pre-training with {KITTI} dataset,
only daylight scenes from the Mini version of {nuScenes} dataset are
chosen for the inference. A sample is composed of 6 images and a 360Â°
{LiDAR} scan. However, in our network, each image is used separately,
and for each image, only the corresponding 3D points inside its
frustum are kept. In other words, each initial scene with a 3D {LiDAR}
scan is split into 6 different scenes corresponding to each camera.

Table~\ref{table:nuscenes_ap} presents the \mbox{AP} results for the
application of several configurations of the proposed network and
PointPillars pre-trained on {KITTI} and applied on {nuScenes}. Because
{nuScenes} is considered here as test data, the point cloud at
inference time is not altered.  Scenes in {nuScenes} tend to be more
crowded than the {KITTI} ones, so the \num{40} top proposals are
extracted.

\begin{table}[H]
    \caption{nuScenes \mbox{AP} for {IoU} $0.5$.}%
  \begin{center}
    \begin{tabular}{c | c | c | c}
      \toprule
      \mbox{Method} & \mbox{Car} & \mbox{Pedestrian} & \mbox{Cyclist} \\
      \midrule
      $PP\_all$ & $19.35$ & $0.49$ & $0.04$ \\
      $PP\_few$ & $35.22$ & $8.69$ & $1.12$ \\
      
      $3D\_all$ & $18.71$ & $4.07$ & $0.07$ \\
      $3D\_few$ & $40.05$ & $12.96$ & $3.08$ \\
      
      $Cam3D\_all$ & $25.75$ & $3.32$ & $0.85$ \\
      $Cam3D\_few$ & $53.65$ & $20.95$ & $2.73$ \\
      
      $Cam\_all$ & $21.63$ & $6.19$ & $1.66$ \\
      $Cam\_few$ & $\bm{57.14}$ & $\bm{21.38}$ & $\bm{8.66}$ \\
      \bottomrule
    \end{tabular}%
    \label{table:nuscenes_ap}
  \end{center}
\end{table}

As expected, \mbox{AP} values decreased drastically with
data that are far from the training distribution. This observation is noticeable on all
experiments but especially on {PointPillars} $PP\_all$. For point
cloud-based experiments $PP\_$ and $3D\_$ the results heavily depend
only on input point clouds. Even if the
data augmentation improves the final results, pedestrians and cyclists remains the main
weaknesses. The different layout of the point cloud layers, due to the
different sensor, and the different environments make small targets
more difficult to detect.  However, $Cam\_$ experiments resist better
to this change of context. Generally, the targets keep the same
overall aspect, even when captured with different camera sensors with
same operation mode (no fish-eye for example) and on the same lighting
conditions. Hence, {RGB} features tend to vary less between
datasets allowing to retrieve small targets even if few 3D data is
available. Nevertheless, the final results are still affected by the
layout of the voxels in the {BEV} map that depends on the input point
cloud.

Furthermore, all the previous findings are still valid in the
{nuScenes} case: the experiments labeled $\_few$ perform better than
the ones trained on unaltered data. In fact, by altering the point
cloud topology during the training by means of the
data augmentation procedure proposed in this paper, the network does
not focus on a specific type of layout but better adapt to variations
in input data. In that case, the change of {LiDAR} sensor between
{KITTI} and {nuScenes} datasets at inference time is the main cause of
data variation.

Concerning \mbox{Cyclist} class, as the number of labels in the
dataset is minimal, each missed target causes an important drop in
\mbox{AP} scores. Moreover, on each data type, \mbox{Cyclist}
class may be badly estimated. {nuScenes} point clouds allocate fewer
points to objects close to the ground, thus it is more difficult to
identify a bicycle of a bike. On {RGB} images, depending on its apparent
pose, the target can be easily confused with a pedestrian. However,
this kind of error will be considered as a false positive on the
Pedestrian class and a false negative on \mbox{Cyclist} class,
causing in the end an important drop in performance.

\begin{figure*}[htb]
  \centering
  \subfloat{\includegraphics[width=0.24\linewidth]{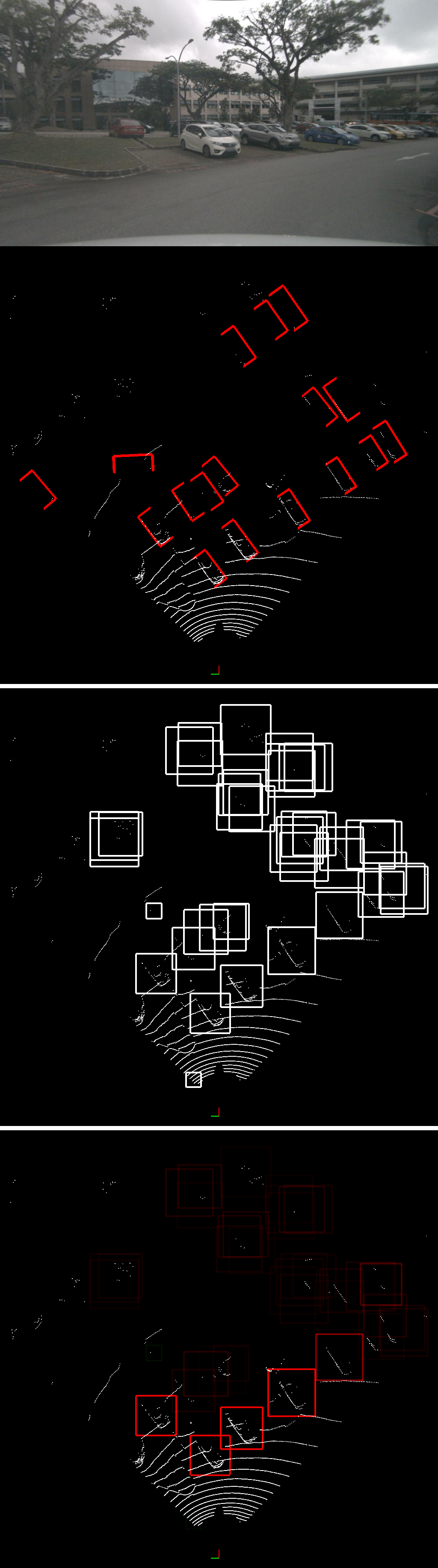}}
  \hfill
  \subfloat{\includegraphics[width=0.24\linewidth]{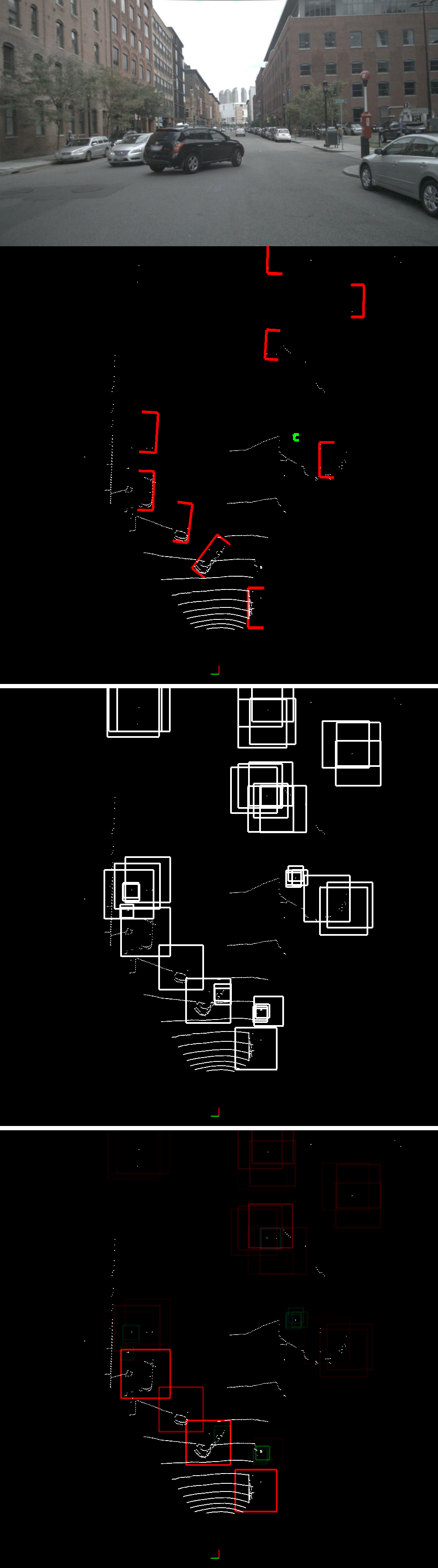}}
  \hfill
  \subfloat{\includegraphics[width=0.24\linewidth]{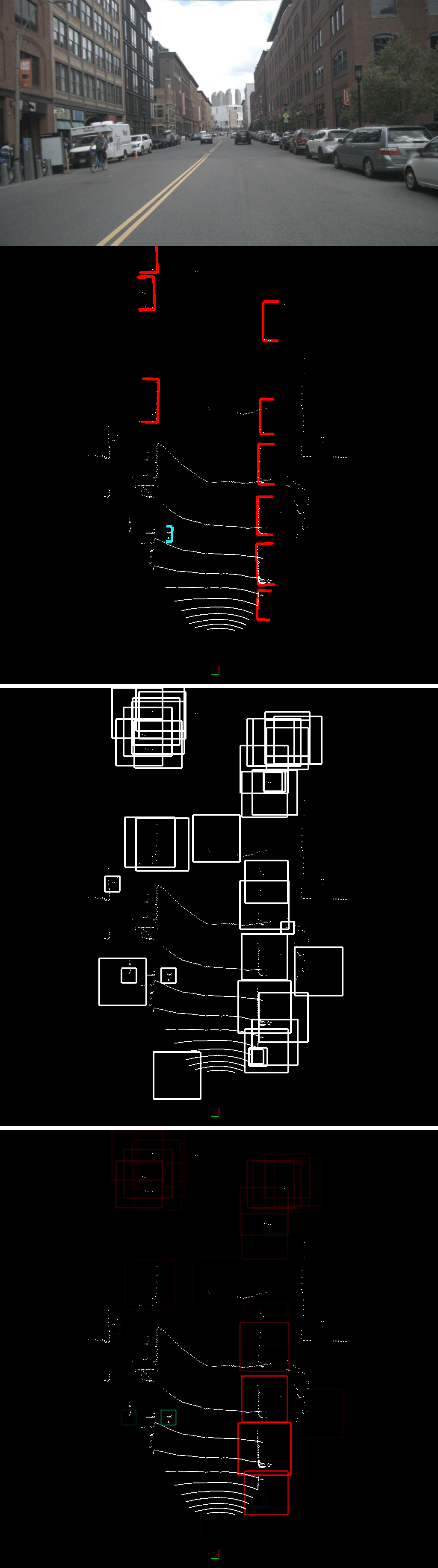}}
  \hfill
  \subfloat{\includegraphics[width=0.24\linewidth]{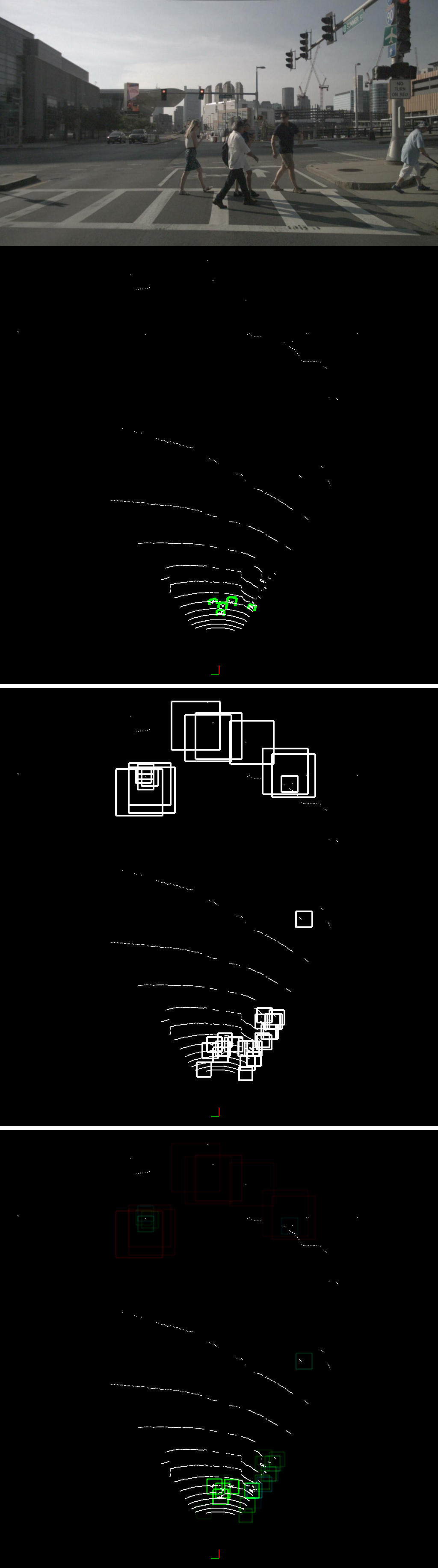}}
  \caption{Qualitative results on {nuScenes}. The first row contains
    input camera images. The second one indicates the labels (red:
    \mbox{Car}, green: \mbox{Pedestrian}, cyan: \mbox{Cyclist}). The
    third row exposes the 40 top proposals without the score delivered
    by the network nor the estimated class. The fourth row illustrates
    the proposals with their estimated class and score through the box
    color and transparency.}%
  \label{fig:nuscenes_images} 
\end{figure*}

Figure~\ref{fig:nuscenes_images} illustrates some visual results given
by $Cam\_few$. It can be noticed that globally, the proposals cover
most of the ground truth boxes. The developed method aims region
proposal but not refined box detection. In that way, we displayed in
the third line the 40 main proposals. Overall, the provided regions of
interest follow the arrangement of the ground truths (second
line). However, the fourth line shows that the confidence scores on
the proposals are often not well estimated causing the drop in
\mbox{AP}. Even if the global aspect of the targets is recognizable,
many parameters can cause the confidence score to decrease on the new
dataset, for example car models never seen in the {KITTI} dataset, the
contrast between the foreground targets and the background, and even
the arrangement of voxels in the {BEV} map.

This experiment illustrates that even without additional training, our
network resists well to change of data domains. The {RGB}-based
network reacts better than the tested point-cloud based methods
because of the smaller variations in the input data, images remaining
regular grids of pixels and convolutions allowing the application of
the same operators that are independent of the camera sensor.

\section{Conclusion}

This paper proposes {R-AGNO-RPN}, a Region Proposal Network that takes
advantage of fusion data of Camera images and {LiDAR} point cloud to
deliver accurate axis-aligned bounding boxes, even if the 3D point cloud is scarcer and
sparser. Feature maps from 2D images are sampled and then scattered on
a Bird's Eye View map thanks to the input point cloud. This fusion of
2D and 3D information is then used to estimate the category, the
location and the dimensions of regions of interest that encompass
possible obstacles. Moreover, we introduced a layer-based data
augmentation technique in order to enforce the {RGB} network to learn
more discriminating features at the first blocks to improve the final
performance.  Our network is compared to a
state-of-the-art method, {PointPillars}, to
illustrate its ability to provide accurate results on high and low
resolution {LiDAR}s. Experiments involving only point cloud resolution
show that the proposed network is able to deliver accurate predictions
even if the input point cloud contains few layers. The performances on
64 layers is also improved thanks to the training procedure that
forces each pixel to become more informative on its own instead of
relying only on its neighborhood.  Finally, the experiments on the
{nuScenes} dataset exposed promising results on the application of the
proposed network on data issued from different sensors ({LiDAR} and
{Camera}) and recorded on different environments. In fact, despite
being only trained on the {KITTI} dataset, the network is able to
locate the potential targets even with the different sensor data
existing in the {nuScenes} dataset. Since the presented system only
focuses on region proposition, future work may concern the addition of
the second stage to accurately regress the bounding boxes such as
precise orientation and the refined dimension and the domain
adaptation for better portability on various urban data.

\section*{Acknowledgments}%
This work has been sponsored by Sherpa Engineering and ANRT
(Association Nationale de la Recherche et de la Technologie). This
work has been sponsored also by the French government research Program
Investissements d'Avenir through the RobotEx Equipment of Excellence
(ANR-10-EQPX-44) and the IMobS3 Laboratory of Excellence
(ANR-10-LABX-16-01), by the European Union through the program
Regional competitiveness and employment 2014-2020 (FEDER - AURA
region) and by the AURA region. We are grateful to the M\'esocentre
Clermont Auvergne University and/or AuBi platform for providing help
and/or computing and/or storage resources.

\bibliographystyle{IEEEtran}
\bibliography{bibliography}
\end{document}